\documentclass[final,1p,times,authoryear]{elsarticle}
\usepackage{geometry}
\usepackage[acronym]{glossaries}
\usepackage{float}
\usepackage{subcaption} 
\usepackage{hyperref}
\usepackage{cleveref}
\usepackage{pgffor}
\usepackage{booktabs}
\usepackage{amsmath}

\journal{Neuroimage}

\makeglossaries

\newacronym{pet}{PET}{Positron Emission Tomography}
\newacronym{vae}{VAE}{Variational Autoencoder}
\newacronym{adas13}{ADAS13}{Alzheimer's Disease Assesment Scale}
\newacronym{glm}{GLM}{General Linear Model}
\newacronym{rsn}{RSN}{Resting State Network}
\newacronym{ndd}{NDDs}{Neurodegenerative Diseases}
\newacronym{adni}{ADNI}{Alzheimer's Disease Neuroimaging Initiative}
\newacronym{ad}{AD}{Alzheimer's Disease}
\newacronym{fdg}{FDG-PET}{Fluorodeoxyglucose Positron Emission Tomography}
\newacronym{fmri}{fMRI}{Functional Magnetic Resonance Imaging}
\newacronym{smri}{sMRI}{Structural Magnetic Resonance Imaging}
\newacronym{mri}{MRI}{Magnetic Resonance Imaging}
\newacronym{elbo}{ELBO}{Evidence Lower Bound}
\newacronym{kl}{KL}{Kullback-Leibler}
\newacronym{dmn}{DMN}{Default Mode Network}
\newacronym{cen}{CEN}{Central Executive Network}
\newacronym{fpn}{FPN}{Fronto-Parietal Network}
\newacronym{mse}{MSE}{Mean Square Error}
\newacronym{mmd}{MMD}{Maximum Mean Discrepancy}
\newacronym{ssim}{SSIM}{Structural Similarity Index}
\newacronym{bce}{BCE}{Binary Cross-Entropy}
\newacronym{hc}{HC}{Healthy Controls}
\newacronym{fpns}{FPNs}{Fronto-Parietal Networks}
\newacronym{mci}{MCI}{Mild Cognitive Impairment}

\begin{document}
	
	\thispagestyle{empty}
	
	\vspace*{3cm}
	
	\begin{center}
		{\LARGE \textbf{IMPORTANT NOTICE}}\\[1.5cm]
		
		{\large
			This manuscript has been peer-reviewed and published in its final form in \textit{NeuroImage}.
		}\\[1.5cm]
		
		{\large
			\textbf{Please cite the official version:}
		}\\[0.5cm]
		
		{\large
			\textbf{An Explainable Framework for the Relationship between Dementia and Metabolism Patterns}
		}\\[0.5cm]
		
		{\large
			DOI: \href{https://doi.org/10.1016/j.neuroimage.2026.121855}{10.1016/j.neuroimage.2026.121855}
		}\\[1.5cm]
		
		{\normalsize
			Please cite the published version instead of this preprint.
		}
	\end{center}
	
	\vfill
	\newpage
	
	
	\begin{frontmatter}

		\title{An Explainable Framework for the Relationship between Dementia and Metabolism Patterns}

		\author{
			C. Vázquez-García\textsuperscript{a,CA}, 
			F. J. Martínez-Murcia\textsuperscript{a, CA}, 
			F. Segovia\textsuperscript{a}, 
			A. Forte\textsuperscript{b},
			J. Ramírez\textsuperscript{a},
			I. Illán\textsuperscript{a},
			A. Hernández-Segura\textsuperscript{a},
			C. Jiménez-Mesa\textsuperscript{c},
			J. M. Górriz \textsuperscript{a,}\textsuperscript{1},
			for the Alzheimer's Disease Neuroimaging Initiative\textsuperscript{*}
		}
		
		\address{\textsuperscript{a}Department of Signal Processing and Biomedical Applications, University of Granada, Granada 18071, Spain. Emails: \{crisvgarcia, fjesusmartinez, fsegovia, javierrp, illan, alejandrohs, gorriz\}@ugr.es\\
			\textsuperscript{b}Department of Statistics and Operations Research, University of Valencia, Valencia 46010, Spain. Email: anabel.forte@uv.es \\
			\textsuperscript{c} Department of Communication Engineering, University of Malaga 29071, Spain. Email: carmenj@uma.es}

		\fntext[J.M.]{J.M. Gorriz is an Associate Editor for NeuroImage and the Executive Guest Editor for the 'Advancing Multimodal Neuroimaging: Explainable and Responsible AI for Early Dementia Detection' Special Issue and was not involved in the editorial review or the decision to publish this article.}

		\tnotetext[ADNI]{Data used in preparation of this article were obtained from the Alzheimer's Disease Neuroimaging Initiative (ADNI) database (adni.loni.usc.edu). As such, the investigators within the ADNI contributed to the design and implementation of ADNI and/or provided data but did not participate in the analysis or writing of this report. A complete listing of ADNI investigators can be found at: \url{http://adni.loni.usc.edu/wp-content/uploads/how_to_apply/ADNI_Acknowledgement_List.pdf}\newline			
		}

		\begin{abstract}
			High-dimensional neuroimaging data poses a challenge for the clinical assessment of neurodegenerative diseases, as it involves complex non-linear relationships that are difficult to disentangle using traditional methods. Variational Autoencoders (VAEs) provide a powerful framework for encoding neuroimaging scans into lower-dimensional latent spaces that capture meaningful disease-related features. In this work, we propose a semi-supervised \acrshort{vae} framework that incorporates a flexible similarity regularization term designed to align selected latent variables with clinical or biomarker measures related to dementia progression. This approach allows adapting the similarity metric and the supervised variables according to specific goals or available data. We demonstrate the framework using \acrfull{pet} scans from the \acrfull{adni} database, guiding the model to capture neurodegenerative patterns associated with \acrfull{ad} by maximizing the similarity between the first latent dimension with a clinical cognitive score, and the second dimension with age. Leveraging the first supervised latent variable, we generate average reconstructions corresponding to different levels of cognitive impairment. A voxel-wise \acrfull{glm} confirms reduced metabolism in key brain regions, predominantly in the hippocampus, and within major \acrfull{rsn}s, particularly the \acrfull{dmn} and the \acrfull{cen}. Further examination of the remaining latent variables show that they encode affine transformations\textemdash rotation, translation, and scaling\textemdash as well as intensity variations, capturing common confounding factors such as inter-subject variability and site-related noise. Our findings indicate that the framework effectively disentangles this neuroimaging biomarker ($z_0$) from confounding factors and age, providing an interpretable and adaptable tool to model and visualize neurodegenerative progression.
		\end{abstract}
		
		

	\end{frontmatter}
	
	\section{Introduction}
	\label{sec1}
	
	\label{sec:introduction}
	
	\acrfull{ndd}, such as \acrshort{ad}, are characterized by progressive brain atrophy and cognitive decline. The global increase in life expectancy is contributing to a rising prevalence of \acrshort{ndd}, with \acrshort{ad} and other dementias showing a $168.7\%$ increase between 1990 and 2021 \cite{steinmetz2024global}. These diseases profoundly affect individuals’ autonomy and quality of life, while also placing significant emotional, social, and financial burdens on caregivers and healthcare systems. Consequently, there is an urgent need for effective tools that enable prognosis and treatment.
	
	Neuroimaging provides valuable insights into the neural alterations underlying these disorders; however, extracting meaningful patterns from its high-dimensional structural and functional data remains a complex task typically requiring expert interpretation. This limitation hampers early intervention, as \acrshort{ad} is often diagnosed only when the damage is already substantial and treatment options are limited \cite{jack2010hypothetical}. By identifying subtle alterations in brain structure and function, imaging techniques can help characterize disease patterns, track progression trajectories, and potentially distinguish high-risk individuals at preclinical stages \cite{frisoni2010clinical}.
	
	Given these challenges, dimensionality reduction techniques have become essential for the analysis of neuroimaging data. This high-dimensionality makes it difficult to identify disease-related patterns, as relevant information is often embedded within complex, non-linear relationships.  The \textit{manifold hypothesis} \cite{tenenbaum2000global, cunningham2014dimensionality, bengio2013representation} suggests that, despite the apparent high-dimensionality of the neuroimaging data, meaningful representations lie on a lower dimensional manifold. By mapping neuroimaging data to a structured latent space, we can uncover disease-related patterns that might otherwise remain hidden in raw imaging data.
	
	In this context, numerous studies have explored both linear and non-linear approaches to uncover latent representations in neuroimaging data, aiming to better characterize neurodegenerative processes. Early works based on linear generative models, such as \cite{dukart2013generative}, used \acrshort{glm}-based frameworks to describe atrophy and hypometabolism in \acrfull{fdg} and \acrfull{smri} data, providing valuable insights despite inherent limitations in modeling complex non-linear patterns.
	
	To address the complex and non-linear data, we use \acrshort{vae}s, a type of generative model that offers significant advantages \cite{kingma2013auto}. \acrshort{vae}s allow us to learn representations of data, which we can map back into the brain space. This capability enables the analysis of how specific latent patterns correspond to variations in brain structures, making it a powerful tool for studying neurodegeneration patterns. Additionally, \acrshort{vae}s are particularly suited for this task due to their explicit nature, which allows direct access to the variables of the learned latent distribution \cite{dalca2019unsupervised}, facilitating the analysis of potential relationships between latent representations and clinical biomarkers.
	
	The standard \acrshort{vae} framework assumes a generative model $p(x,z)$ that captures the joint distribution between the observed data $X$ and a hidden space $Z$. The \acrshort{vae} learns to map from $X$ onto a latent manifold, where smooth transitions along the manifold correspond to smooth changes in the learned latent variables, which encode subject-specific characteristics. The encoder learns an approximate posterior that maps each observation to a distribution over latent variables. As a result, the model organizes the latent space such that similar data points are placed nearby, and traversing specific directions in the manifold leads to interpretable smooth variations in the latent features.
	
	Several works have explored the application of \acrshort{vae}s to neuroimaging data for dementia research. \cite{wakefield2024variational} applied a graph-based unsupervised \acrshort{vae} to \acrshort{fdg} scans, achieving high accuracy in distinguishing \acrshort{ad} patients from controls while enhancing explainability through latent representations. Similarly, \cite{dolci2024interpretable} built a classification model using both genetic and neuroimaging data, extracting features through a \acrshort{vae}. Their model achieved high accuracy and unraveled key regions for the classification task. In a related approach, \cite{kumaranormative} proposed a multimodal \acrshort{vae} for normative modeling of brain \acrshort{mri}, successfully capturing deviations associated with disease progression.
	
	Despite these advances, there remains a need for neuroimaging-derived biomarkers that can capture disease severity in an interpretable manner. Clinical scores provide quantitative measures of cognitive impairment, but they do not directly reveal which brain regions or networks are affected. Conversely, traditional neuroimaging analysis can identify affected regions but often lacks a continuous biomarker that parallels clinical progression.
	
	In this work, we address this gap by proposing a semi-supervised \acrshort{vae} framework designed to learn a latent neuroimaging biomarker that is explicitly aligned with dementia symptomatology. Our primary objective is not classification, but rather to:
	
	\begin{enumerate}
		\item Derive an interpretable neuroimaging biomarker ($z_0$) that encodes dementia severity and correlates with established clinical measures;
		\item Disentangle this biomarker from confounding factors (spatial transformations, intensity variations, inter-subject variability, age) encoded in remaining latent variables;
		\item Leverage the generative capabilities of the model to map this biomarker back to brain space, enabling visualization of metabolic patterns associated with disease progression.
	\end{enumerate}
	
	The main contributions of this work are: First, a semi-supervised VAE framework that learns a neuroimaging biomarker explicitly guided to capture dementia severity through a similarity regularization term. Second, we provide an analysis of the convergence behavior of the model, identifying conditions that prevent both mean collapse and the collapse of regularization terms. Third, we leverage the generative capabilities of the model to map neurodegenerative patterns into brain space, yielding voxel-wise representations that affect how dementia information is encoded in the latent space. Finally, we design an interpretable mechanism to represent common confounding factors \textemdash such as intersubject variability and acquisition noise\textemdash within the model.

	
	\section{Materials and Methods}
	
	\subsection{Data description}
	
	Data used in the preparation of this article were obtained from the \acrshort{adni} database (adni.loni.usc.edu). The \acrshort{adni} was launched in 2003 as a public-private partnership, led by Principal Investigator Michael W. Weiner, MD. The original goal of ADNI was to test whether serial \acrshort{mri}, \acrshort{pet}, other biological markers, and clinical and neuropsychological assessment can be combined to measure the progression of mild cognitive impairment (MCI) and early \acrshort{ad}. The current goals include validating biomarkers for clinical trials, improving the generalizability of \acrshort{adni} data by increasing diversity in the participant cohort, and to provide data concerning the diagnosis and progression of \acrshort{ad} to the scientific community. For up-to-date information, see adni.loni.usc.edu.
	
	We selected a subset containing 3466 \acrshort{fdg} scans, which measure brain glucose metabolism, a key indicator of neurodegeneration \cite{mosconi2005brain}. In addition to imaging data, we included structural biomarker volumes (Hippocampus, Medial temporal lobes, Entorhinal cortex, and Fusiform), normalized by brain size to reduce intersubject variability. Since multiple longitudinal scans per subject are available in \acrshort{adni}, dataset partitioning was performed strictly at the subject level to prevent data leakage. Specifically, unique subject identifiers (PTID) were first extracted, randomly shuffled, and then assigned to training, validation, and test sets according to the predefined proportions: $65\%$ (n = 2239), $15\%$ (n = 572) and $20\%$ (n = 655) respectively. All scans corresponding to a given subject were allocated exclusively to the same subset. Table \ref{training_val_test} summarizes the number of subjects and scans included in each subset.
	\begin{table}[ht]
		\centering
		\begin{tabular}{lccccc}
			\toprule
			& \# Subjects & \# Scans & Age (mean $\pm$ std) & Male & Female \\
			\midrule
			Training   & 1060 & 2239 & 72.8 $\pm$ 7.3 & 593 (55.9$\%$) & 467 (44.1$\%$) \\
			Validation & 244  & 572 & 73.8 $\pm$ 7.0 & 138 (56.6$\%$) & 106 (43.4$\%$) \\
			Test       & 328  & 655 & 73.8 $\pm$ 7.2 & 175 (53.4$\%$) & 153 (46.6$\%$) \\
			\bottomrule
		\end{tabular}
		\caption{Demographics details and number of subjects and scans in each split.}
		\label{training_val_test}
	\end{table} 
	
	\subsection{Data preprocessing}
		\acrshort{pet} scans were coregistered to a common template using rigid-body transformation (SPM12 \cite{SPM12}), but not spatially normalized, in order to preserve individual anatomical variability for the convolutional architecture. Intensity normalization was conducted through a two-step process: min-max normalization (relative to the 99th percentile), followed by exponential transformations to enhance contrast. We found this normalization pipeline to be the best performing and informative about relevant brain structures, specially in the cortex. Since missing values are rare in the dataset, entries with incomplete data were simply excluded. For training, only age and \acrshort{adas13} values were required alongside scans; therefore, scans with missing \acrshort{adas13} values were removed. 
		
	\subsection{Model description}
	
	A \acrshort{vae} consists of two components: an encoder network, which approximates the posterior distribution $q(z|x)$, mapping the input data $x$ into a latent representation $z$; and a decoder network, which reconstructs the data through the likelihood model $p(x|z)$.
	
	Training is performed by maximizing the marginal log-likelihood $\log p(x)$ of the data, which is generally intractable. Instead, the model optimizes a variational lower bound \cite{kingma2013auto}, known as the \acrfull{elbo}. The \acrshort{elbo} can be written as:
	\begin{equation}\label{elbo}
		\mathcal{L}(x) =
		\underbrace{\mathbb{E}_{z\sim q(z|x)}[p(x|z)]}_{\mathcal{L}_{\text{recon}}} -
		\underbrace{D_{KL}(q(z|x) \| p(z))}_{\mathcal{L}_{\mathrm{KL}}} =
		\mathcal{L}_{\text{recon}} - \mathcal{L}_{\mathrm{KL}},
	\end{equation}
	where the first term $\mathcal{L}_{recon}$ corresponds to the reconstruction accuracy, i.e., how well the decoder can reconstruct the input data from the latent representation, and the second term is the \acrfull{kl} divergence between the approximate posterior and the prior $p(z)$, which regularizes the latent space to follow the prior distribution.
	
	For the purpose of the experiments, we opted to use the \acrfull{mse} as the regularization loss, given its suitability for reconstructing continuous-valued neuroimaging data:
	\begin{equation}
		\mathcal{L}_{MSE} = \frac{1}{N}\sum_{i=1}^N ||x_i - \hat{x_i}||^2.
	\end{equation}
	However, a different choice of reconstruction error\textemdash \acrfull{ssim}, \acrfull{bce}, etc.\textemdash is also valid, as long as it is appropriate for the type of data handled. On the other hand, we chose a specific expression for the \acrshort{kl} divergence, by choosing gaussian formulas for both prior $p(z)$ and posterior $q(z|x)$:
	\begin{equation}
		D_{KL}(q(z|x) \| p(z)) = \frac{1}{2} \sum_{i=1}^{d} \left( 1 + \log(\sigma_i^2) - \mu_i^2 - \sigma_i^2 \right),
	\end{equation}
	where $d$ is the latent space dimensionality, and $\mu_i$ and $\sigma_i^2$ are the mean and variance of the $i$-th variable. Additionally, we employ the Beta-VAE \cite{higgins2017beta}, which introduces an extension to the original ELBO by incorporating a weighting parameter $\beta$ to the divergence loss, such that the regularization of the latent space, $\beta\mathcal{L}_{\mathrm{KL}}$, is now controlled by this parameter. Note that the choice of the KL divergence is not unique, and that other divergences can be used\textemdash the \acrfull{mmd}, for instance \cite{zhao2017infovae}.
	
	In our framework, in addition to the traditional ELBO objective, we also incorporated an additional third loss that encourages that a set\textemdash one or more variables\textemdash of the latent space variables capture relevant patterns of cognitive impairment. We denote this term as $\mathcal{L}_{\text{similarity}}$, which measures the statistical similarity between the set of latent variables and the external variable (dementia score, region volume, etc). The choice of metric to quantify similarity is flexible and can be adapted for the case: Pearson correlation, mutual information, or other functions are valid options. Likewise, the external variable used to compute the similarity can be chosen to be different types of biomarkers, clinical scores or other relevant metrics. Formally, we define this regularization term as:
	\begin{equation}\label{similarity_reg}
		\mathcal{L}_{\text{similarity}} = \mathcal{D}(z_{(k)}, y),
	\end{equation}
	where $z_{(k)}$ is the set of supervised variables, $y$ is the external variable of interest, and $\mathcal{D(\cdot,\cdot)}$ is a general similarity metric. In our particular experimentation, we considered the Pearson correlation between some latent variable $z_j$ and a clinical feature $y$, which results in the following expression for the similarity loss term:
	\begin{equation}\label{pearson_formula}
		\mathcal{L}^{(j)}_{\mathrm{similarity}} = - \frac{
			\sum_{i=1}^N \left(z_j^i - \bar{z}_j\right) \left(y_i - \bar{y}\right)
		}{
			\sqrt{\sum_{i=1}^N \left(z_j^i - \bar{z}_j\right)^2} \,
			\sqrt{\sum_{i=1}^N \left(y_i - \bar{y}\right)^2} 
		} \equiv -r(z_j, y), 
	\end{equation}
	where $N$ is the number of subjects, $z_j^i$ denotes the $j$-th latent variable of the $i$-th subject, $y_i$ is the value of the clinical feature (\acrshort{adas13}, age, etc.) for the $i$-th subject, and $\bar{z}$ and $\bar{y}$ are the mean values of the $j$-th latent variable, and the clinical feature, respectively.
	
	The final total loss function used during training was:
	\begin{equation}\label{eq:loss_total}
		\mathcal{L} = 
		\mathcal{L}_{MSE}
		- \beta\mathcal{L}_{KL} 
		+ \sum_{j=1}^{\alpha_{D}}\alpha_j\mathcal{L}^{(j)}_{similarity},
	\end{equation}
	where $\alpha_D$ is the number of independent variables that correlate to a clinical feature.
	We added $\alpha_j$ as hyperparameter to control the strength of the similarity regularization for each different feature. In this work, we considered the similarity terms $-r(z_0, \text{ADAS13})$ and $-r(z_1, age)$, where \acrfull{adas13} is a cognitive score used in clinical assessment \cite{kueper2018alzheimer}. 
	
	Importantly, the model is not trained to predict clinical scores directly, but to encourage specific latent dimensions to align with imaging patterns associated with these variables. Thus, while \acrshort{adas13} and age explicitly influence the latent structure through the similarity regularization term, all other associations reported in this study \textemdash including correlations with hippocampal, entorhinal and medial temporal volumes, as well as fusiform \textemdash were not incorporated into the training objective but used for post hoc analyses.

	\subsection{Architecture}
	
	The architecture of the semi-supervised 3D convolutional \acrshort{vae} is as follows:
	\begin{itemize}
		\item\textbf{Encoder:}
		\begin{itemize}
			\item[-] Four convolutional layers with kernel sizes of $11$, $7$, $5$, and $3$ (with $32$, $64$, $128$, and $256$ channels, respectively), progressively decreasing in size, each followed by a ReLU activation function and batch normalization to ensure stable training and improve convergence.
			\item[-] A fully connected network transforms  the output of the last convolutional layer, which has 9216 neurons ($256$ channels $\times 3\times 4\times 3$ spatial dimensions), into a linear feature representation of $256$ neurons, followed by ReLU activation.
			\item[-] Two separate fully connected layers map this representation into a mean space and a log-variance space. These are then reparameterized to obtain the final latent space representation.
		\end{itemize}
		\item\textbf{Decoder:}
		\begin{itemize}
			\item[-] A fully connected network maps the latent space back into a linear feature representation of $4608$ neurons ($128$ channels $\times 3\times 4\times 3$ spatial dimensions), followed by a ReLU activation.
			\item[-] The decoder employs three transposed convolutional layers with increasing kernel sizes of $3$, $4$, and $11$ (with $128$, $64$, and $32$ channels, respectively), each followed by a ReLU activation function and batch normalization to progressively reconstruct the input volume.
		\end{itemize}
	\end{itemize}
	\begin{figure}[H]
		\centering
		\includegraphics[width=1\linewidth]{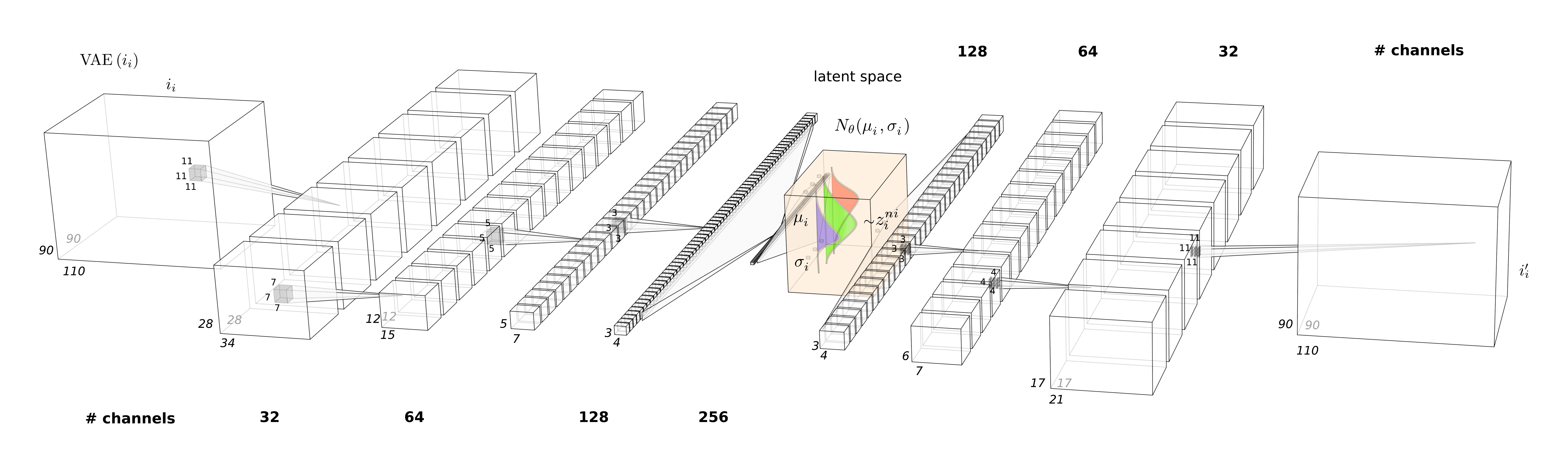}
		\caption{Architecture of the \acrshort{vae} model. Convolutional layers extract features from high dimensional volumes into feature maps, where are hierarchically compressed into lower representations. }
		\label{fig:VAE_architecture}
	\end{figure}
	The model was trained using the PyTorch library, with Adam as the optimizer and a learning rate of $2\times 10^{-5}$. The hyperparameters, including the number of epochs, $\beta$, $\alpha$, batch size, and latent space dimensionality were selected based on preliminary experiments. The dataset was split into training, evaluation, and testing sets with respective proportions of $0.65$, $0.15$, and $0.2$. Batch normalization layers were introduced after each convolutional layer to stabilize training and provide regularization.
	
    In our experiments, correlation was computed using the Pearson correlation $r(z_k, y)$ for all samples in a batch. First, missing values were removed, then data were centered by subtracting the batch mean. The covariance and standard deviations were calculated across the batch, and finally, the Pearson correlation was obtained according to equation \eqref{pearson_formula}. The Pearson loss is defined as the negative of this correlation to maximize alignment between latent dimension $z_k$ and the target $y$.
    	
	\subsection{Model Evaluation and Interpretability}
	
	Our evaluation strategy focuses on validating that $z_0$ functions as a meaningful neuroimaging biomarker aligned with dementia progression. We assess three key aspects:
	
	\begin{enumerate}
		\item Biomarker alignment: Correlation between $z_0$ and both clinical measures and structural biomarkers;
		\item Biological plausibility: Whether the patterns associated with $z_0$ correspond to established AD-affected regions;
		\item Disentanglement: Whether confounding factors are captured by other latent variables rather than $z_0$.
	\end{enumerate}
	
	First, We explored the hyperparameter space to identify the most suitable value ranges for the regularization parameters $\beta$ and $\alpha_j$, as well as the dimensionality of the latent space. To support this analysis, we constructed a set of phase diagrams that illustrate the model's behavior under varying hyperparameter configurations. Phase diagrams\textemdash commonly used in physics and engineering\textemdash serve here to visualize how changes in these parameters affect model performance, by showing how the behavior changes.
	
	In our context, these phase diagrams reveal distinct operational regimes of the \acrshort{vae}, capturing regions where the model successfully learns meaningful latent representations\textemdash which we denote as the \textit{stable regime}\textemdash versus regions where the model collapses to trivial solutions\textemdash denoted by the \textit{failure regime} (e.g., posterior collapse to the mean of the input data). 
	Our phase diagrams clearly delineate the hyperparameter regions where this collapse happens versus those where the model learns useful, informative latent representations. This insight is crucial for tuning \acrshort{vae}s in neuroimaging, where avoiding posterior collapse ensures the latent space captures meaningful biological variability rather than trivial averages.
	
	To quantify this we computed a first phase diagram, where we analyzed the distribution of the mean latent vectors across the validation set. Specifically, we computed the average Euclidean distance of each latent mean vector to the global centroid of all latent means. This metric captures dispersion of the representations in the latent space: higher average distances indicate more diverse and informative encodings, whereas low values suggest that the latent variables have collapsed towards a single point, reflecting a non-informative utilization of the latent space. By studying how this dispersion varies with the latent space dimensionality and the strength of the KL regularization, we identified the stable regimes versus the failure regimes. The computed Euclidean distance metric is given by the formula:
	\begin{equation}\label{phase_diagram_euclidean_dist}
		\mathcal{D_{\mu}} = \frac{1}{N}\sum_{i=1}^N||\mu_i - \overline{\mu}||_2,
	\end{equation}
	where $\overline{\mu} = \frac{1}{N}\sum_{i = 1}^N \mu_i$ is the centroid of all the latent mean vectors $\mu_i$, $||\cdot||_2$ is the euclidean distance, and $N$ is the number of subjects.

	Next, the second phase diagram shows whether the semi-supervised model learns latent representations that are informative about dementia, by varying the hyperparameters $\beta$ and $\alpha_{j=0}$ (in the case of dementia). This analysis explores how the interplay between the KL divergence and the similarity regularization terms affects the model's behavior. Specifically, we compute the Pearson correlation between the first latent dimension $z_0$ and the cognitive score across different $\beta$ and $\alpha$ values to identify the stable regimes\textemdash where model captures meaningful associations with dementia severity\textemdash and failure regimes \textemdash where the similarity is either negligible or overly dominant, leading the model to converge to a non-informative minimum. While we use Pearson correlation in this case, the approach can be generalized to any similarity metric defined in equation \eqref{similarity_reg}. To empirically assess this flexibility, we additionally repeated the training procedure using a rank-based Spearman correlation as similarity metric, suggesting that the results are independent of the chosen similarity metric.
	
	Notice that, while we use the same terminology (stable/failure regimes) for consistency, we emphasize that these regimes reflect different failure modes: either a loss of latent variability due to posterior collapse, or a loss of clinical relevance due to the ineffective or overly strong similarity regularization. Importantly, both conditions must be satisfied for the framework to be considered effective.
	
	To further characterize the learned latent space, we systematically sample from it to generate synthetic subjects and examine the resulting reconstructions. Specifically, for each latent variable $z_l$ (e.g., $z_0$ representing \acrshort{adas13}, and $z_1$ representing age), we generate latent representations by systematically varying those latent variables while sampling the remaining latent dimensions from a standard normal distribution $N(0,1)$ to capture inter-subject variability. These latent codes are then decoded back into brain space and averaged to obtain mean \acrshort{pet} scans, which reflect characteristic brain patterns associated with variations in the considered $z_l$ latent variables.
	
	To quantify the contribution of each latent variable and their potential interactions within a general and extensible framework, we defined a voxel-wise \acrshort{glm} over the reconstructed scans. The voxel intensities of these averaged reconstructions were then used as the dependent variable in the \acrshort{glm}:
		\begin{equation}\label{general_glm_eq}
			\mu(z_0, z_1, \dots, z_{L-1})[i,j,k] = \alpha[i,j,k] + \sum_{l=0}^{L-1}\beta_l[i,j,k]z_l + \sum_{l=0}^{L-1}\sum_{m = l+1}^{L-1}\gamma_{lm}[i,j,k]z_lz_m + \epsilon,
		\end{equation}
	where $L$ denotes the number of latent variables included in the \acrshort{glm}, which may be chosen according to the specific experimental design, $\alpha[i,j,k]$ is the baseline intensity for voxel $(i,j,k)$, $\beta_l[i,j,k]$ captures the effect of latent $z_l$, $\gamma_{lm}[i,j,k]$ models interactions between latent variables $z_l$ and $z_m$, and $\epsilon$ is the residual error. The regression coefficients were estimated using ordinary least squares. The resulting maps of $\beta_l$ coefficients capture voxel-wise variations as a function of each latent dimension conditioned on the rest, while the $\gamma_{lm}$ maps highlight regions where interactions between latent factors contribute to brain activity patterns. While eq. \eqref{general_glm_eq} defines the general formulation of the framework, in the experiments presented in this work we restricted the \acrshort{glm} to the two guided latent dimensions, namely $z_0$ (dementia-related) and $z_1$ (age-related), i.e., $L=2$. The remaining latent variables were not included in the regression analysis as they capture unsupervised variability not explicitly associated with predefined clinical factors.
	
	Additionally, to quantitatively assess the discriminative power of the learned representations, we performed binary logistic regression to classify \acrshort{ad} patients and \acrfull{hc}. The diagnosis labels were binarized, with \acrshort{hc} coded as 0 and \acrshort{ad} as 1.
	
	The latent representations $z=(z_0, z_1, \cdots, z_{k})$ ($k$ being the latent dimensionality) from both training and test sets were extracted and used as input features. Subjects with missing diagnostic information were excluded from the analysis. We trained a logistic regression model with a maximum of 1000 iterations using the training set latent representations as predictors and binary diagnosis as the outcome. The trained model was then applied to test set to generate predictions.
	
	To assess the individual contribution of different components of the latent space, we performed classification using two distinct input configurations: (a) all latent variables $(z_0, \cdots, z_k)$, (b) only the cognitive score \acrshort{adas13} as input. Classification performance was evaluated using accuracy, sensitivity (recall), specificity, and balanced accuracy metrics. Results were visualized through boxplots showing the distribution of each latent variable across diagnostic groups, allowing us to illustrate the differential predictive power of each latent dimension. To ensure robustness of the results, we employed a bootstrap validation \cite{efron1993introduction} approach, repeating the procedure 100 times and reporting the average performance.

	Moreover, to further explore and interpret the latent space, we perform a systematic variation along each individual latent dimension. Starting from a latent vector initialized to zero, we vary one latent variable at a time across its observed range and map the resulting vectors back to brain space. This enables us to visualize how changes in each specific latent variable influence the reconstructed brain images, providing a visual insight into the role of each dimension.
	
	Finally, to assess the contribution of the similarity regularization to the model, we conducted an ablation study in which this term was removed. This enables a direct comparison between the patterns learned with and without the similarity constraint. The results are then directly compared to the full model in order to understand the difference between guided and unguided patterns.


	\section{Results}
	
	\subsection{Hyperparameter Selection and Model Convergence}
	
\begin{figure}[ht]
		\centering
		\includegraphics[width=1.0\textwidth]{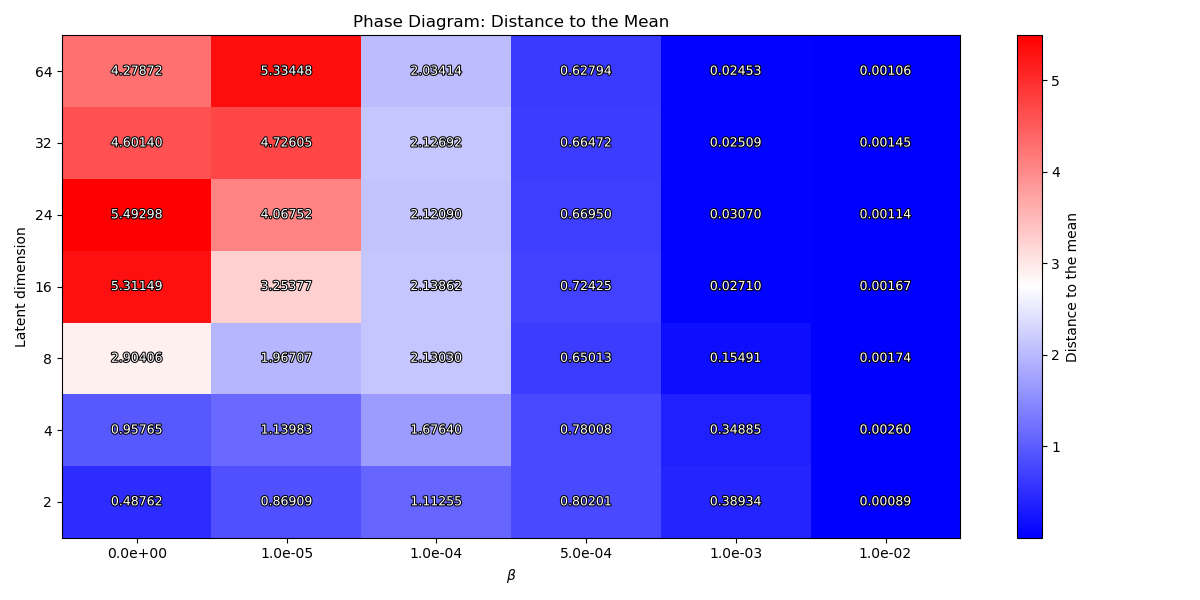}
		\caption{Phase diagram of latent dimensionality vs. $\beta$. The color code shows the mean euclidean distance to the mean $\mathcal{D_{\mu}}$ of the latent variables. Large values of $\beta$ cause the model to collapse to the mean, whereas small values of latent dimension are not sufficient to capture the variability of the data.}
		\label{fig:phase_diagram_conv}
	\end{figure}

\begin{figure}[ht]
		\centering
		\includegraphics[width=1.0\textwidth]{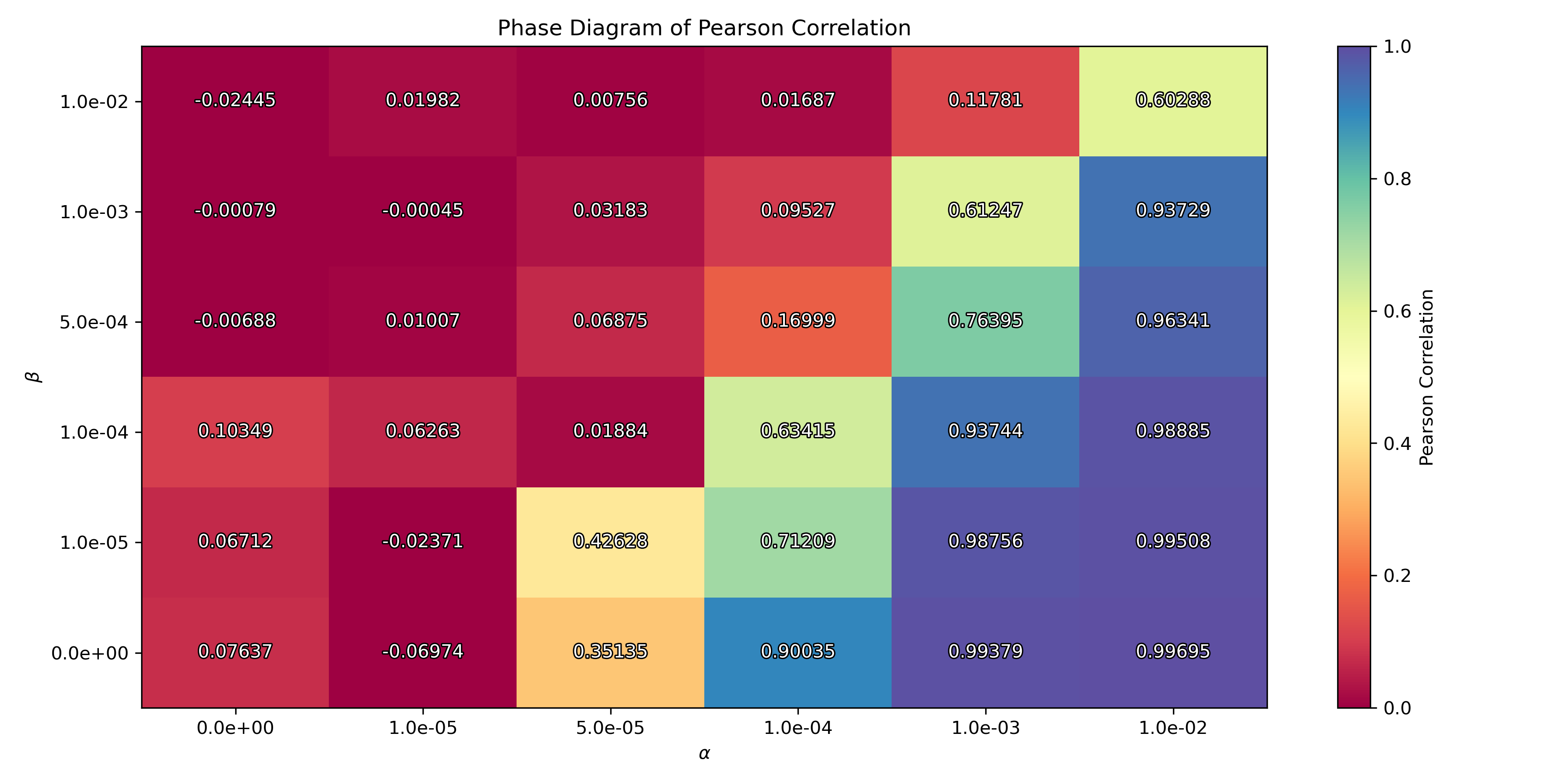}
		\caption{Phase diagram of KL divergence vs. Pearson regularization for $\alpha_{j=0}$. The color code shows the convergence of the Pearson correlation. Large values of $\alpha$ lead to non-informative regularization of the latent space, whereas small values do not produce patterns correlated with dementia. The latent dimensionality is constant at $8$.}
		\label{fig:phase_diagram_corr}
	\end{figure}

	To examine model convergence and parameter ranges, figures \ref{fig:phase_diagram_conv} and \ref{fig:phase_diagram_corr} present the resulting phase diagrams. In the first figure, the color scale represents the mean euclidean distance to the mean of the latent representation across dimensions, computed using equation \eqref{phase_diagram_euclidean_dist}. Here, we fix $\alpha_j = 0,  \forall\, j = 1, \ldots, \alpha_D$ to isolate the effect of KL regularization on the model's convergence behavior.

	Three distinct regions emerge in this first phase diagram, depending on the $\mathcal{D_{\mu}}$ values. First, a deep blue region indicates very low deviation from the mean, suggesting model collapse to the data mean and therefore a failure regime. This region is associated with either high $\beta$ values (strong KL regularization) or small latent dimensions. Second, a red region reflects high variance, indicating well-separated latent representations with large variability. This region indicates that the model acts almost like a deterministic autoencoder, also corresponding to a failure regime. Finally, a third intermediate region, corresponding to light colors transitioning from blue to pink, shows the stable regime, where representations exhibit variability but remain close within the latent space.

	The second phase diagram displays the convergence values of the Pearson correlation between the first latent variable $z_0$ and the cognitive score (\acrshort{adas13}). In this analysis, the coefficients $\alpha_{j\neq0}$, associated with different variables (such as age \textemdash the only other variable included in our experiments) are excluded; although the same procedure could be applied to any variable, our primary focus is the dementia-related biomarker. The diagram shows how this correlation evolves as a function of the regularization parameters $\beta$ and $\alpha_{j=0}$, while keeping the latent dimensionality fixed. We identify three distinct regions. The red region corresponds to near-zero correlation, indicating that the latent space fails to capture clinically relevant information, and thus a failure regime. In contrast, the bluish-purple region shows correlation close to one, suggesting that the similarity regularization is overly dominant, providing non-informative representations, which corresponds to a failure regime. We identify the stable regime in the intermediate transition corresponding to light values from orange to blue, where the model achieves meaningful correlations that align with clinical information.
	
	Now, to conduct our experiments and analyze the resulting representations generated by the model, we selected an appropriate set of hyperparameters. The results presented in this work were obtained using the following set of hyperparameters: a latent dimension of 8, a learning rate of $2\times 10^{-5}$, a batch size of 8, $\beta = 1\times 10^{-4}$, $\alpha_{j=0} = 2\times 10^{-4}$, and $\alpha_{j=1} = 2\times 10^{-4}$. It is worth noting, however, that similar values of these parameters yield comparable results, provided they lie within the stable regimes shown in Figs. \ref{fig:phase_diagram_conv} and \ref{fig:phase_diagram_corr}, due to the flexibility of the model.

\begin{figure}[H]
		\centering
		\includegraphics[width=0.7\textwidth]{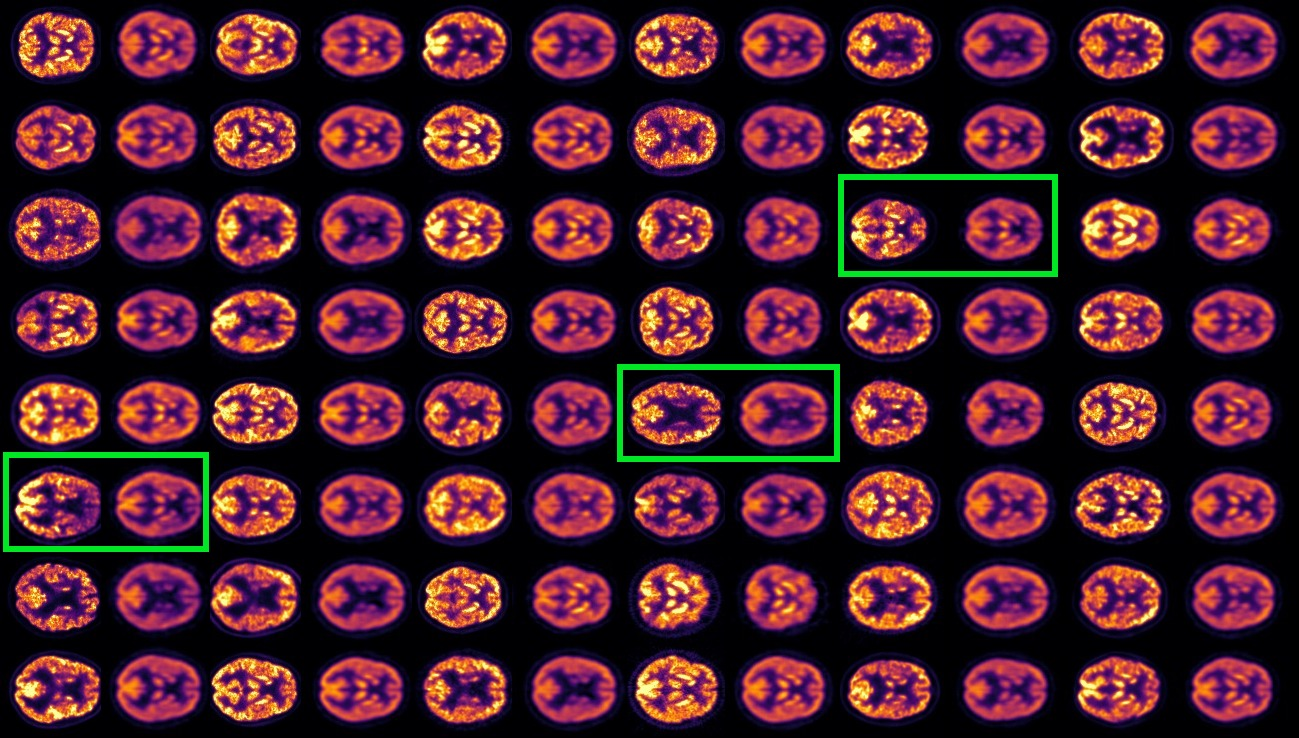}
		\caption{Odd columns: input brain slices. Even columns: VAE reconstructions. Green boxes highlight input-reconstruction pairs. Color represents intensity; reconstructions show lower intensity but preserve relevant structure. The \acrshort{kl} regularization counterweights the reconstruction, producing lower quality scans than the input.}
		\label{fig:reconstruction}
	\end{figure}

	We encoded the input volumes of the test set, obtained their latent codes, and decoded back to brain space, to see the resulting reconstructions, shown in \ref{fig:reconstruction}. In this figure we present input slices (odd columns) along with their corresponding reconstructions (even columns).

	\subsection{$z_0$ as a Neuroimaging Biomarker of Dementia Severity}
	
\begin{figure}[H]
		\centering
		\begin{subfigure}[b]{0.45\textwidth}
			\centering
			\includegraphics[width=\textwidth]{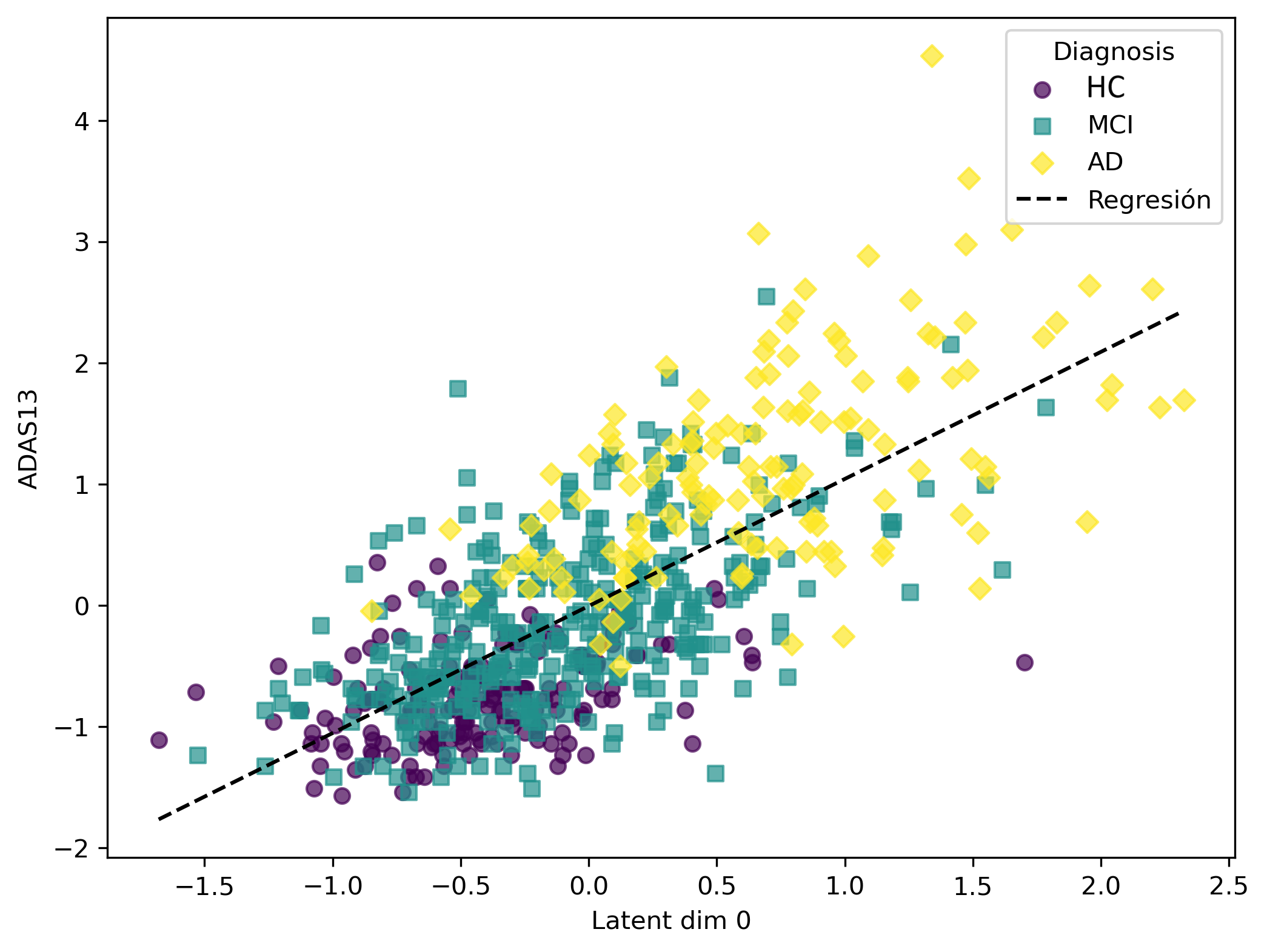}
			\caption{Relationship between the value of the latent variable $z_0$ and the normalized \acrshort{adas13} score. $|r| = 0.790$, $p\ll 0.001$}
			\label{fig:corr_adas}
		\end{subfigure}
		\hfill
		\begin{subfigure}[b]{0.45\textwidth}
			\centering
			\includegraphics[width=\textwidth]{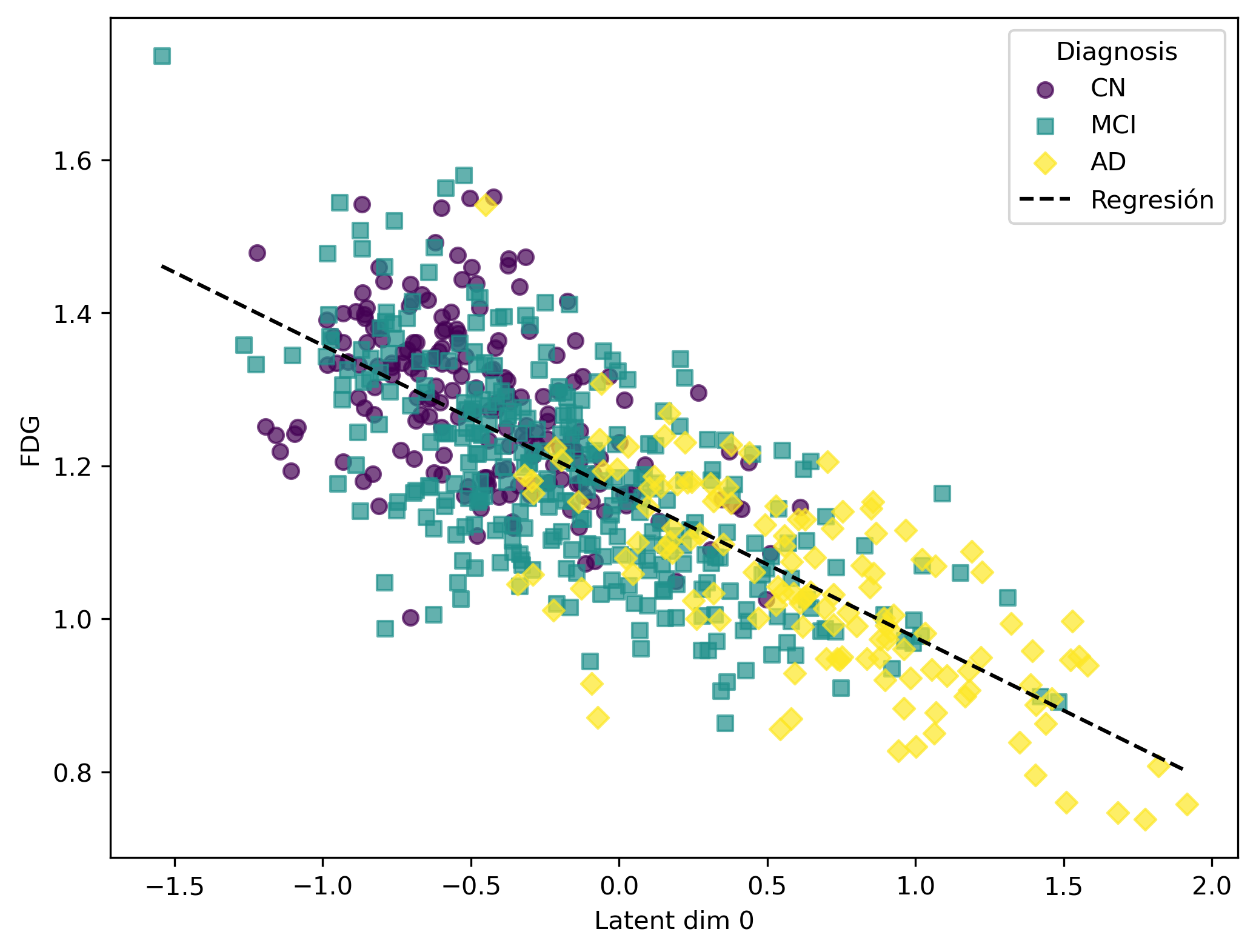}
			\caption{Relationship between the value of the latent variable $z_0$ and the \acrshort{fdg} score. $|r| = 0.810$, $p\ll 0.001$}
			\label{fig:corr_fdg}
		\end{subfigure}
		\caption{Correlation figures between latent variable $z_0$ and (a) the \acrshort{adas13} cognitive score, and (b) the average \acrshort{fdg}. As expected, cognitive decline (\acrshort{adas13}) increases as $z_0$ increases, whereas metabolism (\acrshort{fdg}) decreases as $z_0$ increases.}
		\label{fig:corrs_z0}
	\end{figure}

	The primary objective of our framework is to learn a latent neuroimaging biomarker ($z_0$) that captures dementia progression. Fig. \ref{fig:corrs_z0} demonstrates that $z_0$ successfully arranges subjects along a continuum reflecting cognitive decline ($|r| = 0.790$, $p \ll 0.001$ for \acrshort{adas13}, and $|r| = 0.810$, $p \ll 0.001$ for \acrshort{fdg}), indicating that the model learned a clinically meaningful biomarker of disease severity. The color code shows the diagnosis of the subjects: \acrshort{hc}, \acrfull{mci}, and \acrshort{ad}. 
	
	To further assess the robustness of the similarity regularization with respect to the choice of similarity metric, we retrained the model using the Spearman rank correlation instead of the Pearson correlation within the regularization term. The results were highly consistent with those obtained using Pearson (Spearman correlation $|\rho| = 0.77$, $p \ll 0.001$ for \acrshort{adas13}), demonstrating that the learned association is not dependent on the specific similarity metric employed.
	
\begin{figure}[H]
		\centering
		
		\begin{subfigure}[t]{0.48\columnwidth}
			\centering
			\includegraphics[width=0.8\linewidth]{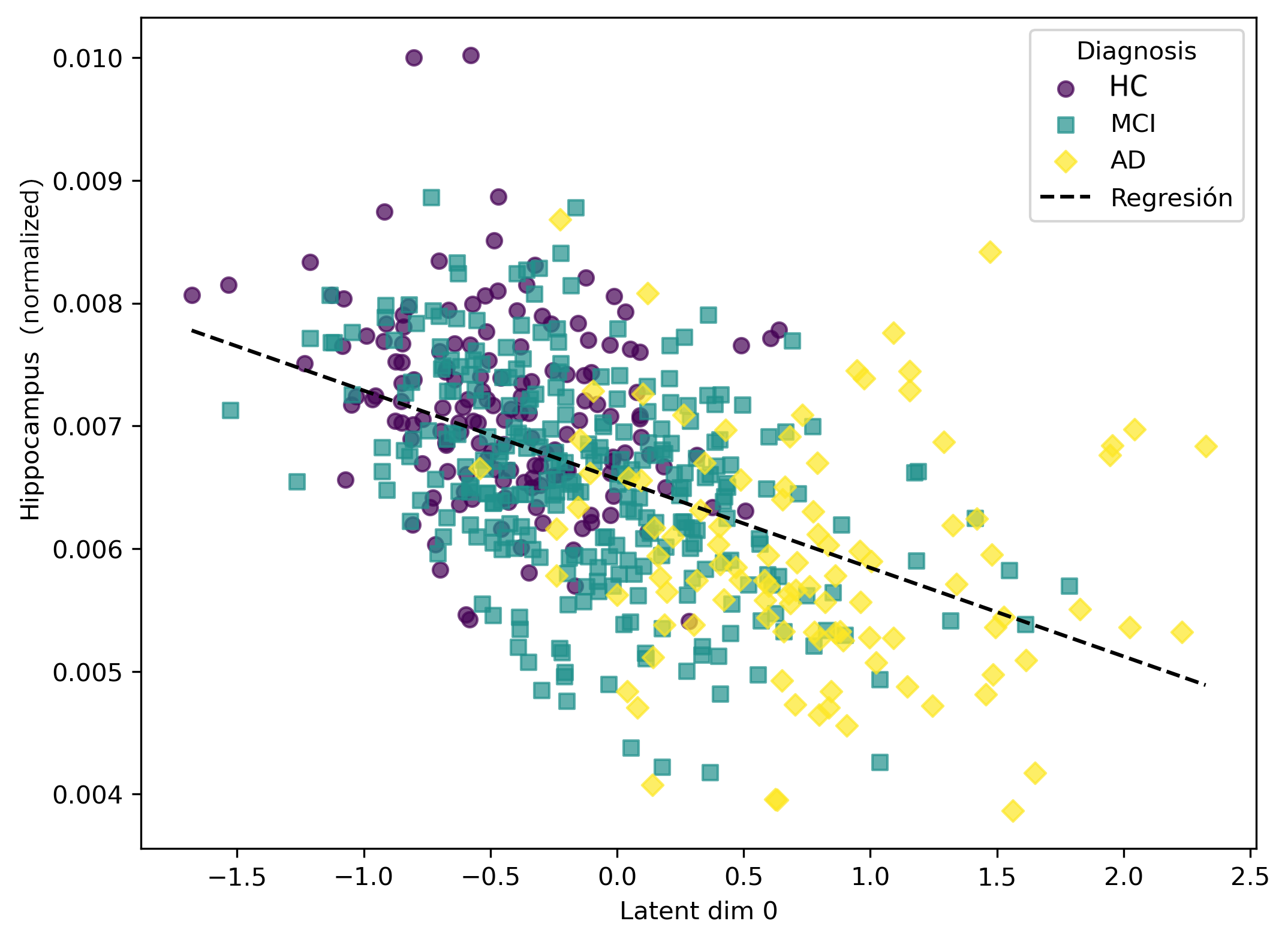}
			\caption{Hippocampal volume. $|r| = 0.48$, $p \ll 0.001$. Color code shows the diagnosis.}
			\label{fig:biomarker_hippocampus}
		\end{subfigure}
		\hfill
		\begin{subfigure}[t]{0.48\columnwidth}
			\centering
			\includegraphics[width=0.8\linewidth]{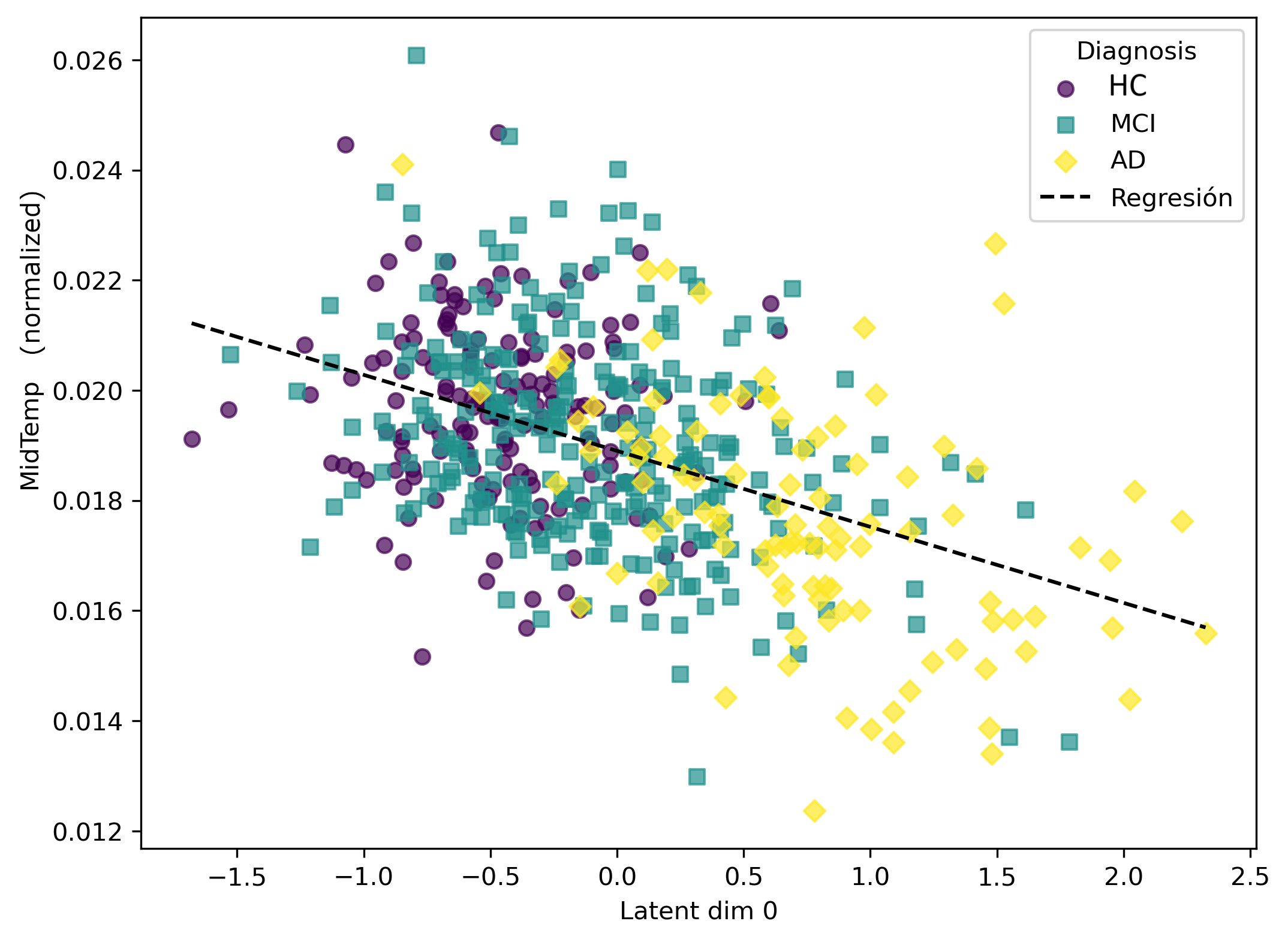}
			\caption{Medial temporal lobes volume. $|r| = 0.45$, $p \ll 0.001$.}
			\label{fig:biomarker_midtemp}
		\end{subfigure}
		
		\vspace{0.5em}
		
		\begin{subfigure}[t]{0.48\columnwidth}
			\centering
			\includegraphics[width=0.8\linewidth]{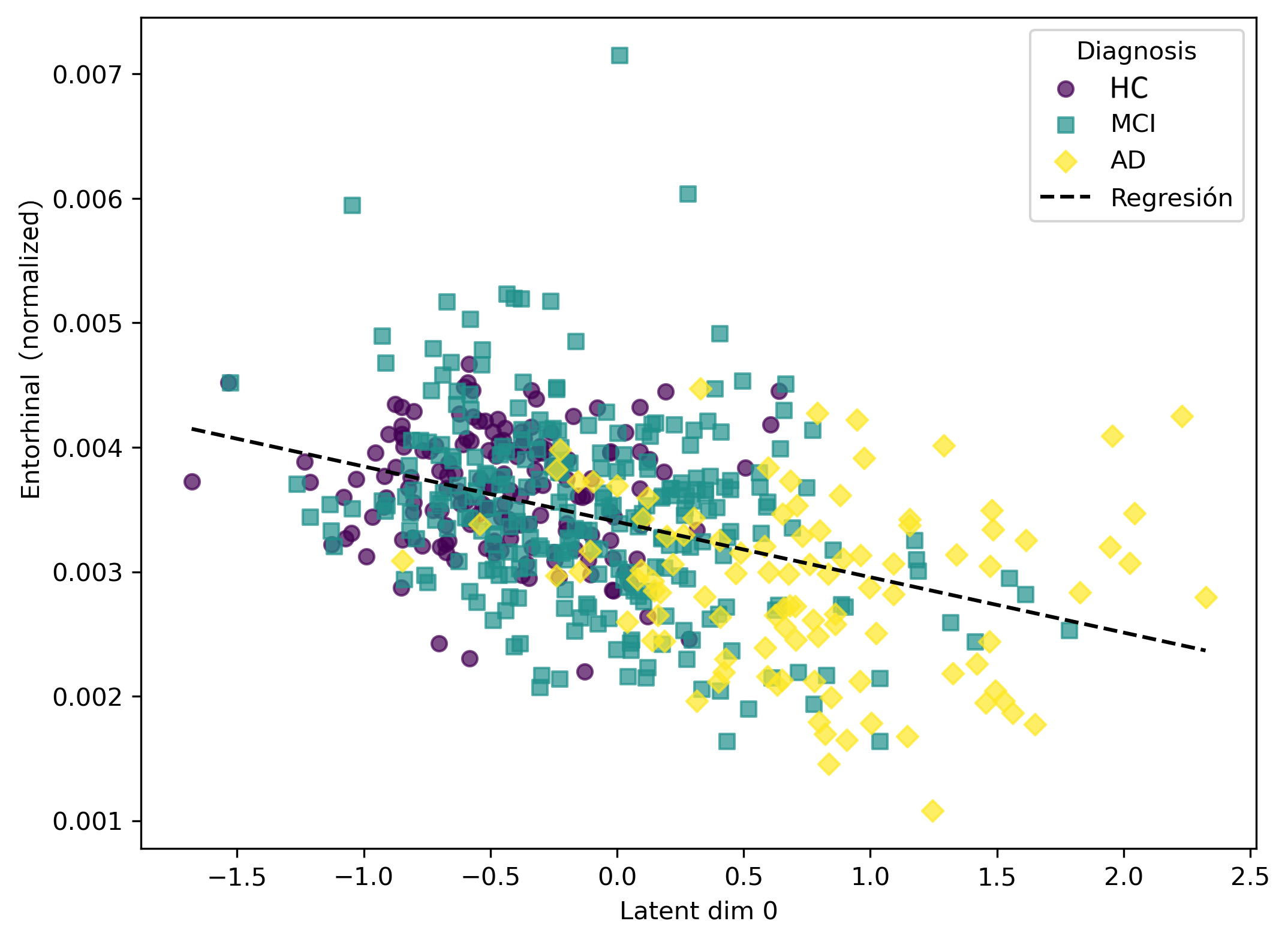}
			\caption{Entorhinal cortex volume. $|r| = 0.37$, $p\ll 0.001$.}
			\label{fig:biomarker_entorhinal}
		\end{subfigure}
		\hfill
		\begin{subfigure}[t]{0.48\columnwidth}
			\centering
			\includegraphics[width=0.8\linewidth]{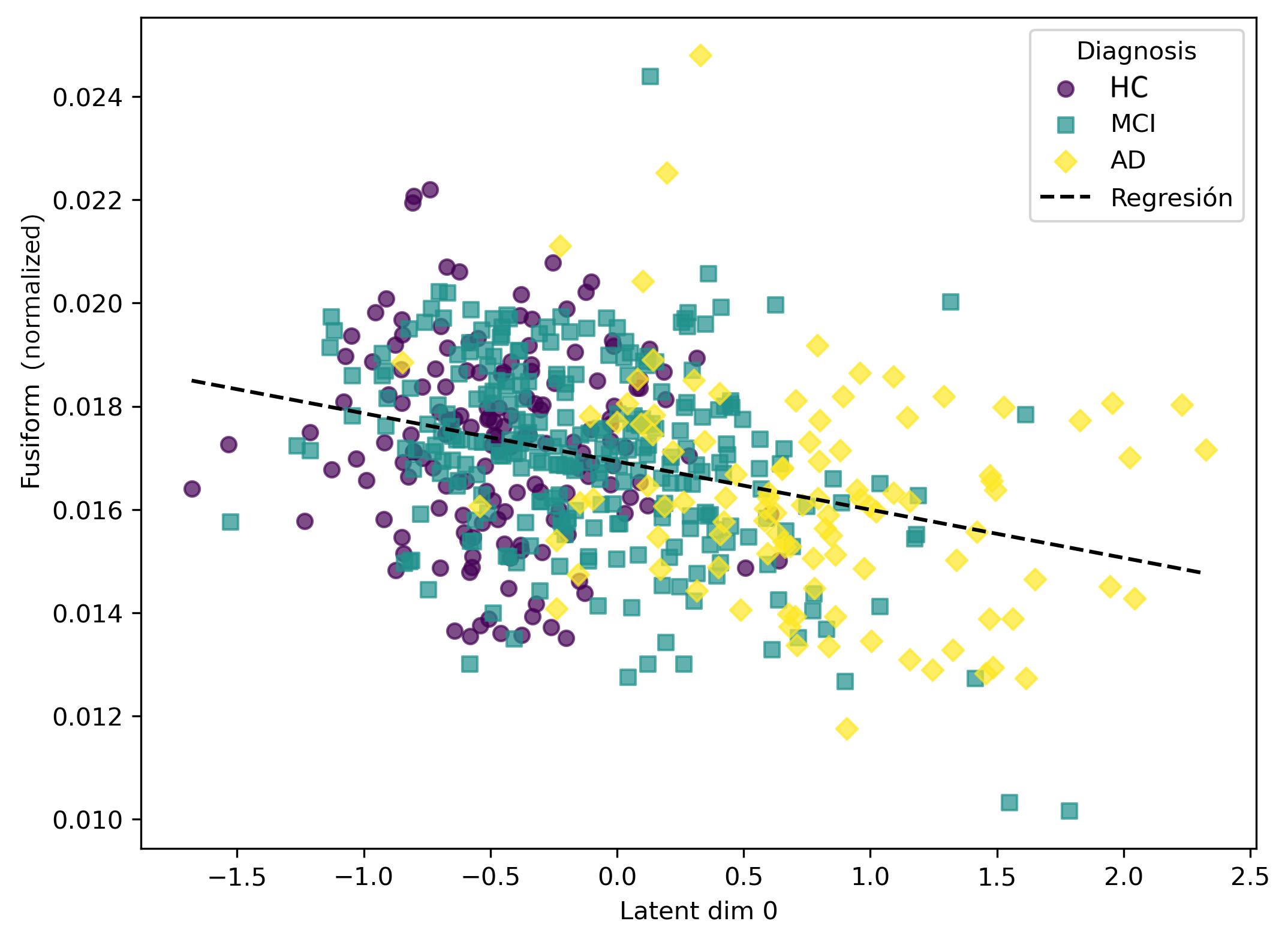}
			\caption{Fusiform volume. $|r| = 0.34$, $p \ll 0.001$.}
			\label{fig:biomarker_fusiform}
		\end{subfigure}
		
		\caption{Correlation between latent representations of test set against the latent variable $z_0$, for several biomarker measures. All volume measures are normalized by the brain size.}
		\label{fig:biomarkers}
	\end{figure}

	To validate the biological relevance of this biomarker, we examined its relationship with established structural biomarkers affected in \acrshort{ad} (Fig. \ref{fig:biomarkers}). $z_0$ showed moderate-to-strong correlations with hippocampal volume (|r|=0.48) \cite{jack2008alzheimer}, medial temporal lobes (|r|=0.45) \cite{wuestefeld2024medial}, and entorhinal cortex (|r|=0.37) \cite{igarashi2023entorhinal}, consistent with the known  progression of \acrshort{ad}. Notably, the hippocampus and medial temporal regions showed robust associations, whereas the fusiform gyrus presented only a weak correlation \cite{braak1991neuropathological}.
	
	\subsection{Visualization of Metabolic Patterns Associated with $z_0$}

\begin{figure}[H]
		\centering
		\begin{subfigure}[b]{0.34\linewidth}
			\includegraphics[width=\linewidth]{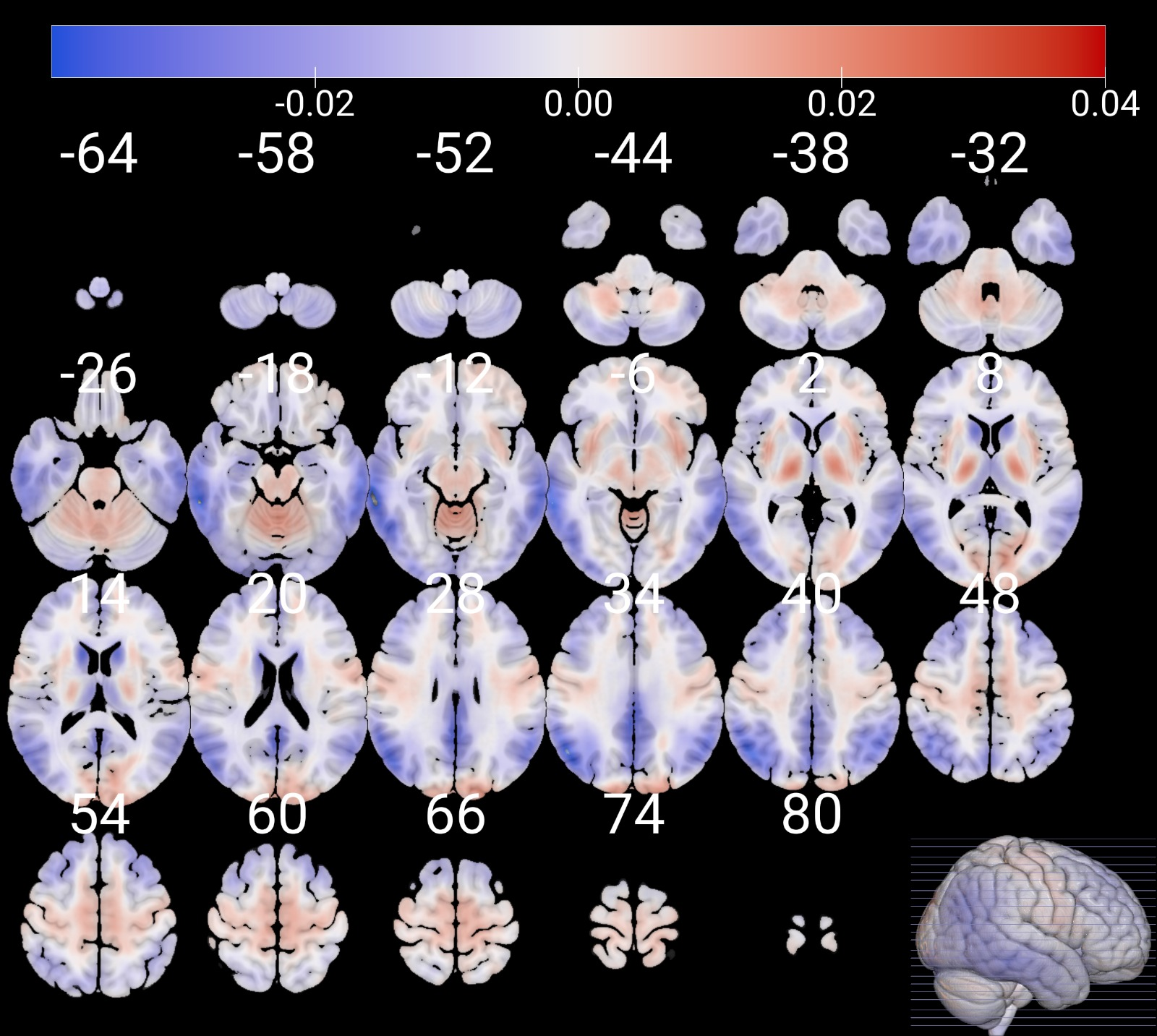}
			\caption{Mosaic axial visualization of the GLM estimation.}
			\label{fig:adas_GLM_voxels}
		\end{subfigure}
		\begin{subfigure}[b]{0.32\linewidth}
			\raisebox{1.5ex}{\includegraphics[width=\linewidth]{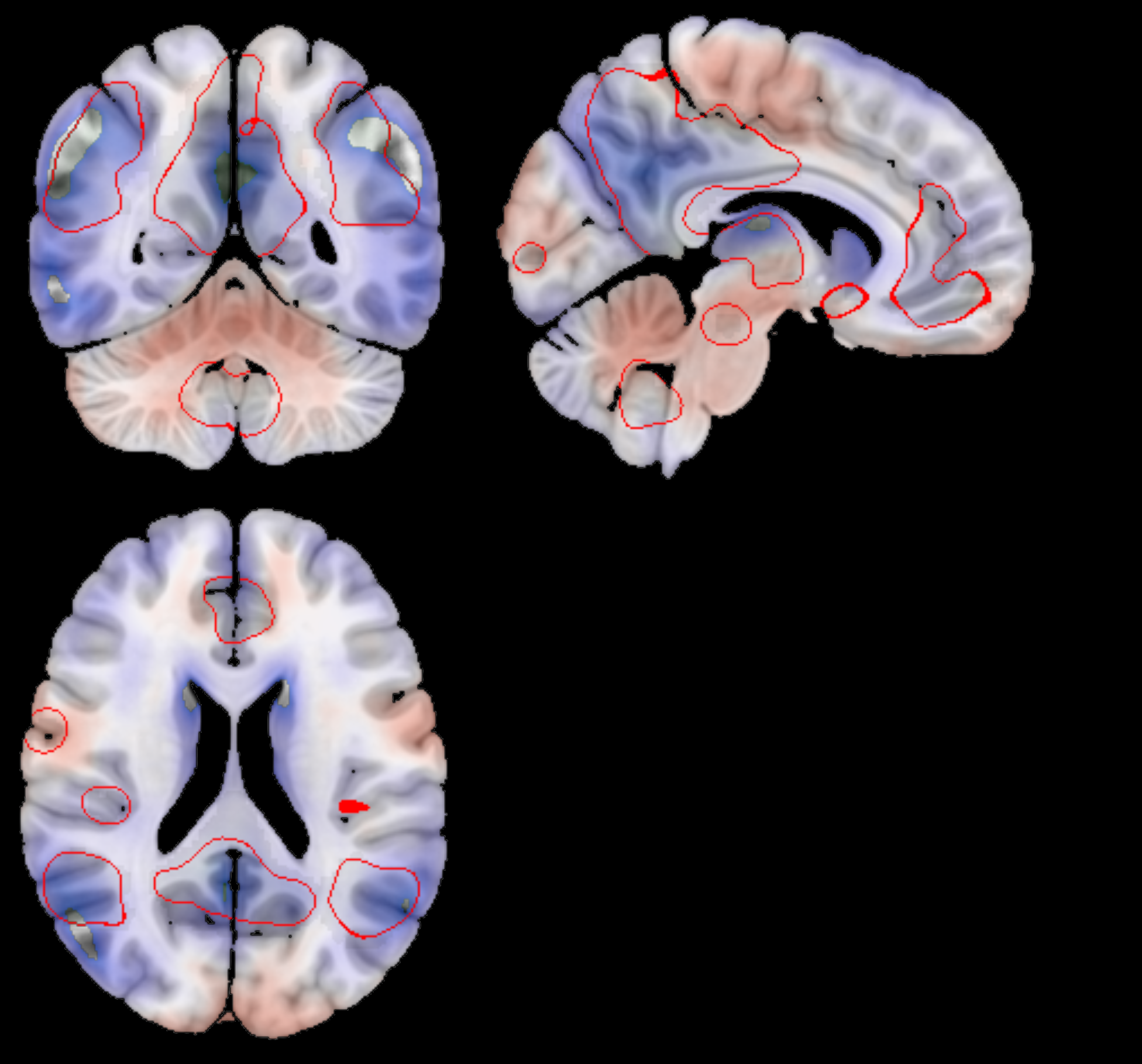}}
			\caption{DMN.}
			\label{fig:adas_GLM_DMN}
		\end{subfigure}
		\begin{subfigure}[b]{0.32\linewidth}
			\raisebox{1.5ex}{\includegraphics[width=\linewidth]{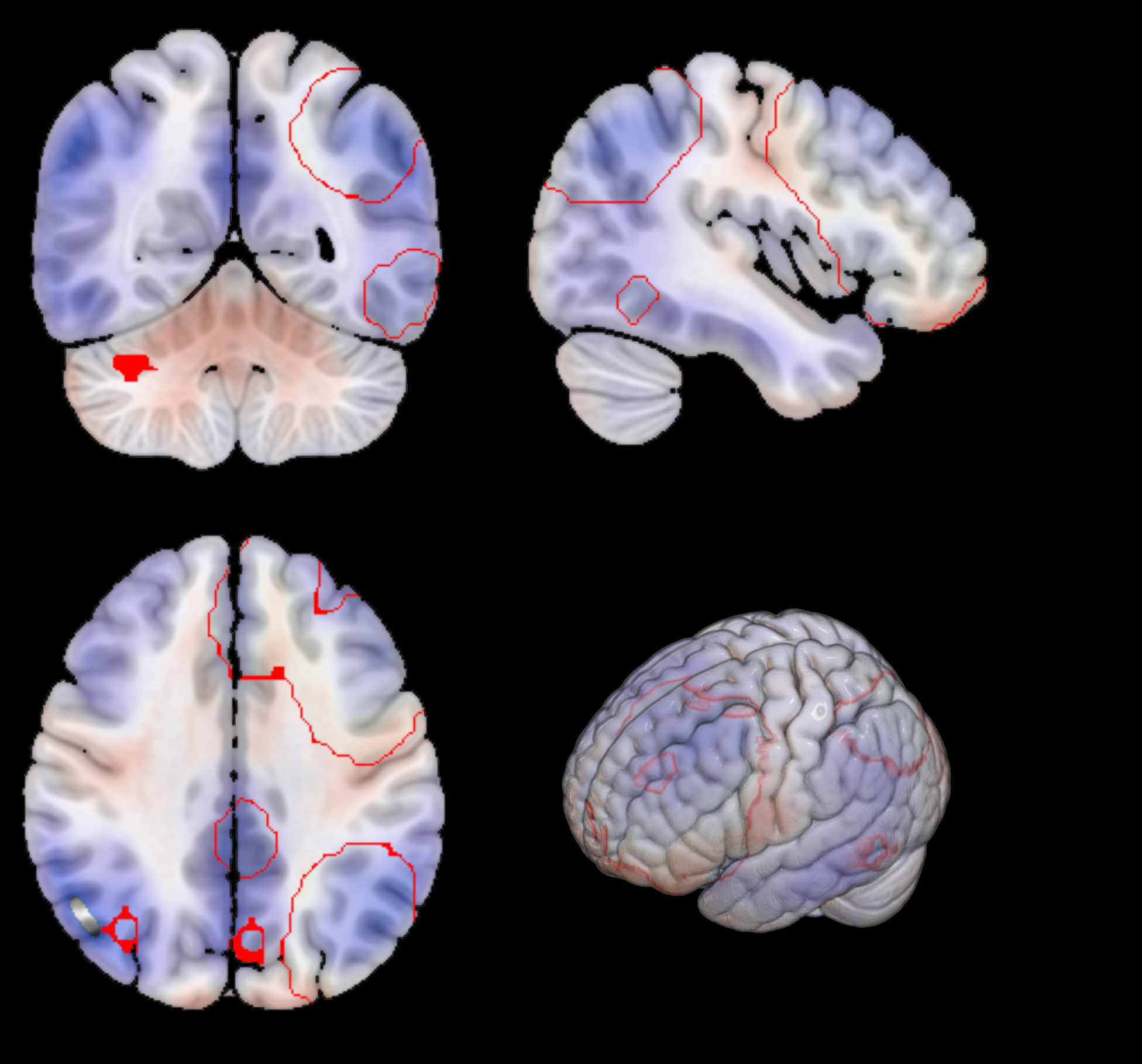}}
			\caption{Left FPN.}
			\label{fig:adas_GLM_FPN_left}
		\end{subfigure}
		\vskip\baselineskip
		\begin{subfigure}[b]{0.34\linewidth}
			\includegraphics[width=\linewidth]{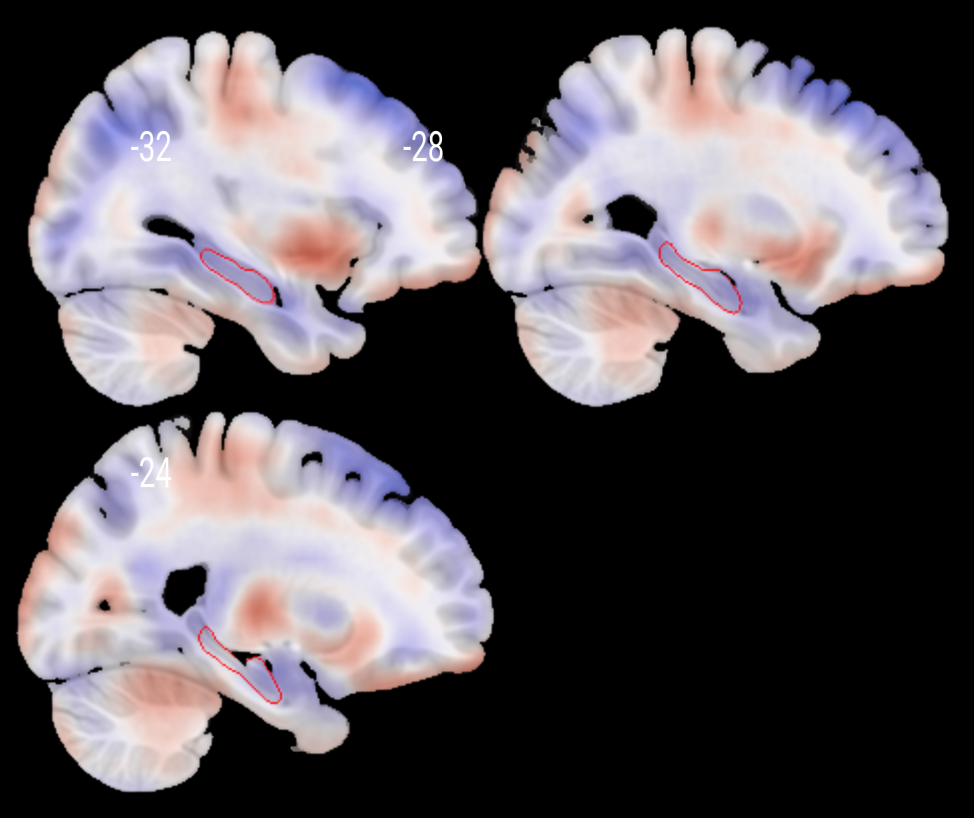}
			\caption{Visualization of the hippocampus. Hippocampus region is highlighted with red outlines.}
			\label{fig:adas_glm_hippocampus}
		\end{subfigure}
		\begin{subfigure}[b]{0.32\linewidth}
			\raisebox{1.5ex}{\includegraphics[width=\linewidth]{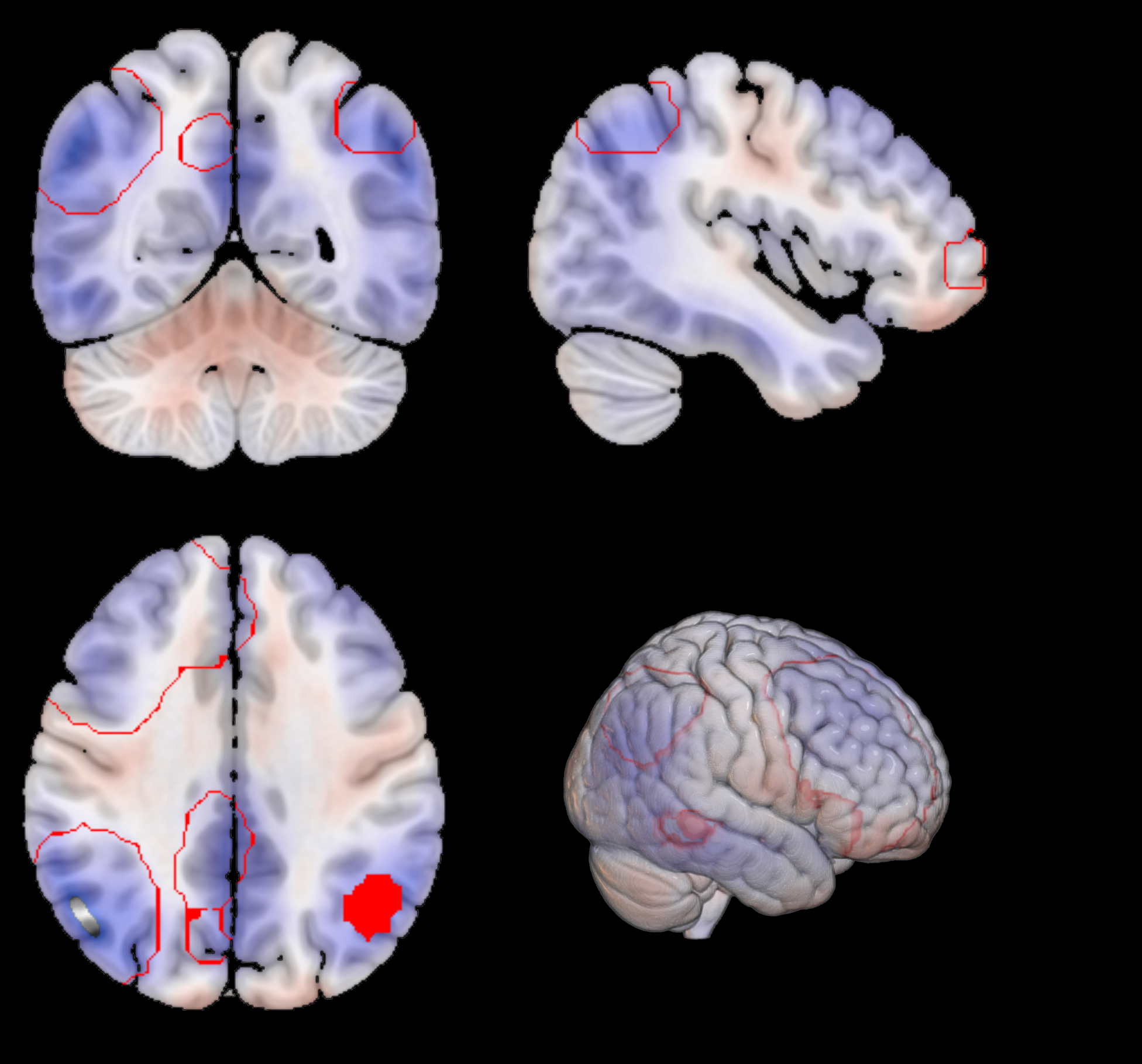}}
			\caption{Right FPN.}
			\label{fig:adas_GLM_FPN_right}
		\end{subfigure}
		\begin{subfigure}[b]{0.32\linewidth}
			\raisebox{1.5ex}{\includegraphics[width=\linewidth]{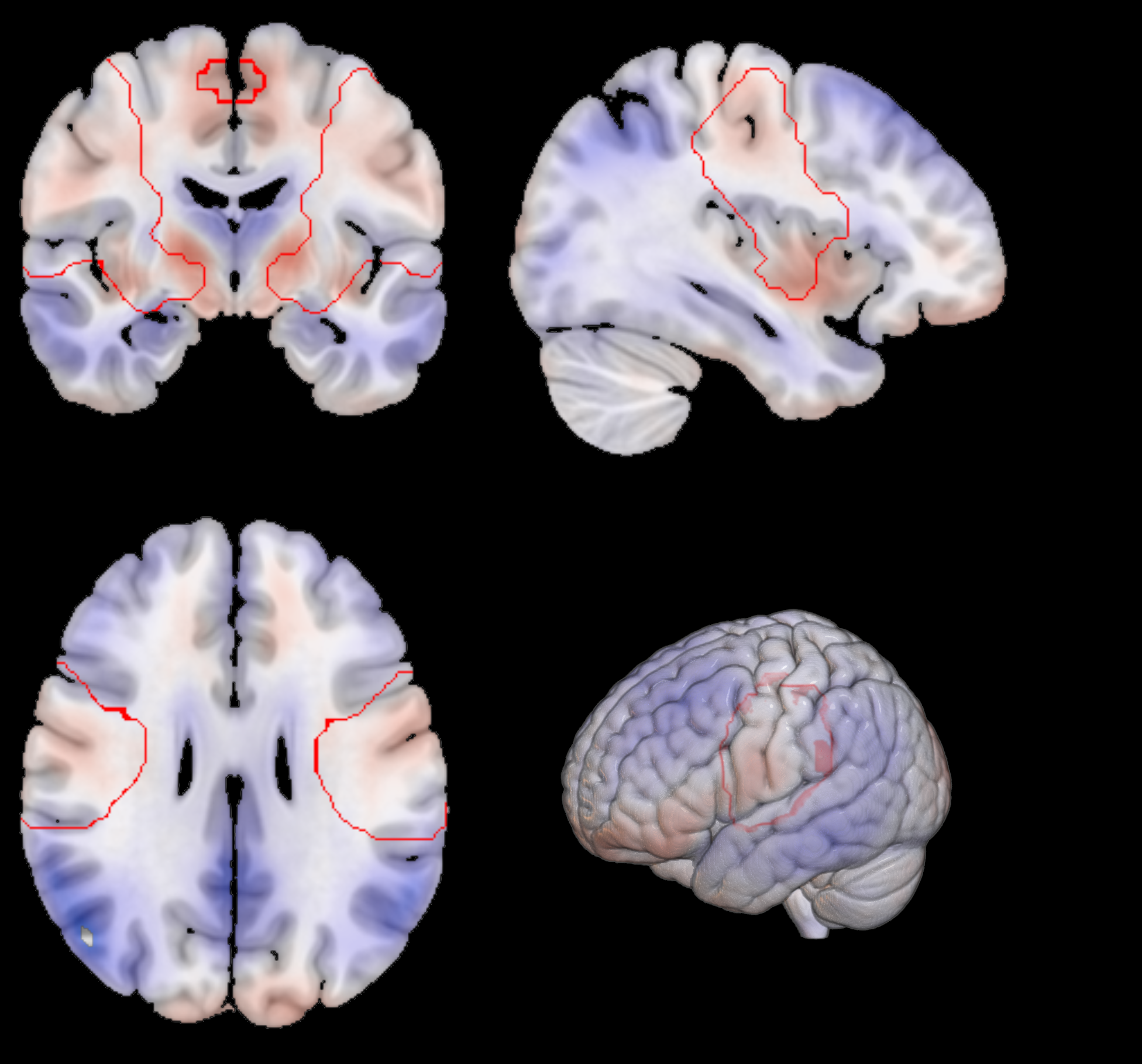}}
			\caption{Sensorimotor Network.}
			\label{fig:adas_glm_sensimotor}
		\end{subfigure}
		\caption{Voxel coefficients for $z_0$ estimated by the GLM model between glucose metabolism and disease severity. Values correspond to the $\beta_0[i,j,k]$ coefficients of the model, illustrating the marginal effect of $z_0$ conditioned to $z_1$ and their interaction. The figure displays the spatial pattern of the effect across the brain. Panel (a) shows a general visualization of the whole brain across different slices. (d) highlights in red the hippocampus ROI across several slices. (b), (c), (e) and (f) highlight some relevant \acrshort{rsn}s. Visualization is performed using the MRIcroGLM sofware \cite{rorden2017mricroglm}.}
		\label{fig:adas_GLM}
	\end{figure}

	Leveraging the generative nature of the \acrshort{vae}, we mapped the neuroimaging biomarker $z_0$ back to brain space to visualize associated metabolic patterns. We generated average latent representations corresponding to fixed values of both $z_0$ (\acrshort{adas13}) and $z_1$ (age) within the observed range (see Fig. \ref{fig:corr_adas}). These representations were decoded into brain space, and a voxel-wise GLM was applied to both latents $z_0$ and $z_1$, according to \eqref{general_glm_eq}, in order to identify metabolism changes. The resulting statistical map for $z_0$ (Fig. \ref{fig:adas_GLM}) highlights regions most affected by neurodegeneration, displaying the voxel-wise $\beta_0[i,j,k]$ coefficients of the GLM rather than thresholded statistical maps. This choice was made to visualize the spatial contribution and relative magnitude of the association between metabolism and disease severity, rather than to assess voxel-wise significance.
	
	Decreased metabolism (in blue) was predominantly observed in the prefrontal and medial temporal cortices, as well as in parts of the occipital lobe. In particular, Fig. \ref{fig:adas_glm_hippocampus} reveals a well-defined decline in metabolic activity in the hippocampal region. In contrast, increased metabolism (in red) was found in the motor cortex (Fig. \ref{fig:adas_glm_sensimotor}) and various subcortical structures, as found in literature \cite{mosconi2005brain, minoshima1997metabolic, biller2016neurology}. It is important to note that these structures were not explicitly provided to the model, but instead emerged post hoc from the latent representations.
	
	Additional insights are provided in Figs. \ref{fig:adas_GLM_DMN}, \ref{fig:adas_GLM_FPN_left}, \ref{fig:adas_GLM_FPN_right}, and \ref{fig:adas_glm_sensimotor}, which show the voxel-wise GLM coefficients overlaid on key \acrshort{rsn}s. These include the \acrshort{dmn} and the left and right \acrfull{fpns}. Both the \acrshort{dmn} and \acrshort{fpn}s demonstrated significant reductions in metabolic activity, whereas the Sensorimotor Network exhibited either no significant changes or slight increases in metabolism.
	
	Although $z_0$ was aligned with \acrshort{adas13} during training, the spatial patterns observed in the \acrshort{glm} coefficients emerge from the data, providing interpretable insights into the neuroanatomical distribution of dementia-related effects.
		
	\subsection{AD classification}
	
	Although classification is not the primary objective of this work, we assessed the discriminative power of the disease-related latent variable $z_0$. This experiment serves to quantify how well this neuroimaging biomarker separates \acrshort{ad} from \acrshort{hc} compared to known benchmarks. The classification results are summarized in Table \ref{tab:classification_report}. Please note that the values for the baseline methods \cite{wakefield2024variational, dolci2024interpretable} are reported directly from the literature and were not re-implemented on the exact data split used in this study. However, inter-study comparability is supported by the fact that all methods utilize \acrshort{fdg} data from the \acrshort{adni} database. While specific subject subsets may vary, potential discrepancies arising from different acquisition protocols (e.g., radiotracer dose, uptake time, or scanner sensitivity) were mitigated through the intensity normalization applied in our preprocessing pipeline \cite{lopez2020intensity}. This ensures that the metabolic patterns remain comparable across different \acrshort{adni} cohorts.
	\begin{table}[ht]
		\centering
		\begin{tabular}{l c c c c}
			\hline
			\textbf{Metric} & Our model & \acrshort{adas13} & \cite{wakefield2024variational} & \cite{dolci2024interpretable} \\
			\hline
			Accuracy               & $0.8 \pm 0.02$ & $0.96\pm0.03$ &  -  & $0.926\pm0.02$ \\
			Sensitivity (recall)   & $0.79 \pm 0.04$ & $0.95\pm0.01$ &-  & $0.876\pm 0.03$ \\
			Specificity            & $0.77 \pm 0.02$ & $0.95\pm0.02$ & -  & - \\
			Balanced Accuracy      & $0.79 \pm 0.02$ & $0.96\pm0.02$ &$0.85 \pm 0.01$ & - \\
			\hline
		\end{tabular}
		\caption{Classification metrics for the \acrshort{ad} vs. \acrshort{hc} task. The values of our model were validated using bootstrap validation with 10 resamples. We also provide a comparison with work from the literature.}
		\label{tab:classification_report}
	\end{table}
	As shown in Table \ref{tab:classification_report}, while dedicated classification frameworks (and the clinical ground truth \acrshort{adas13}) achieve higher accuracy, our model retains significant predictive power ($79\%$ balanced accuracy) using only a single latent dimension ($z_0$). This confirms that $z_0$ successfully captures the core metabolic signal of the disease, disentangled from age and confounding factors.
	
\begin{figure}[ht]
	\centering
	\includegraphics[width=\linewidth, keepaspectratio]{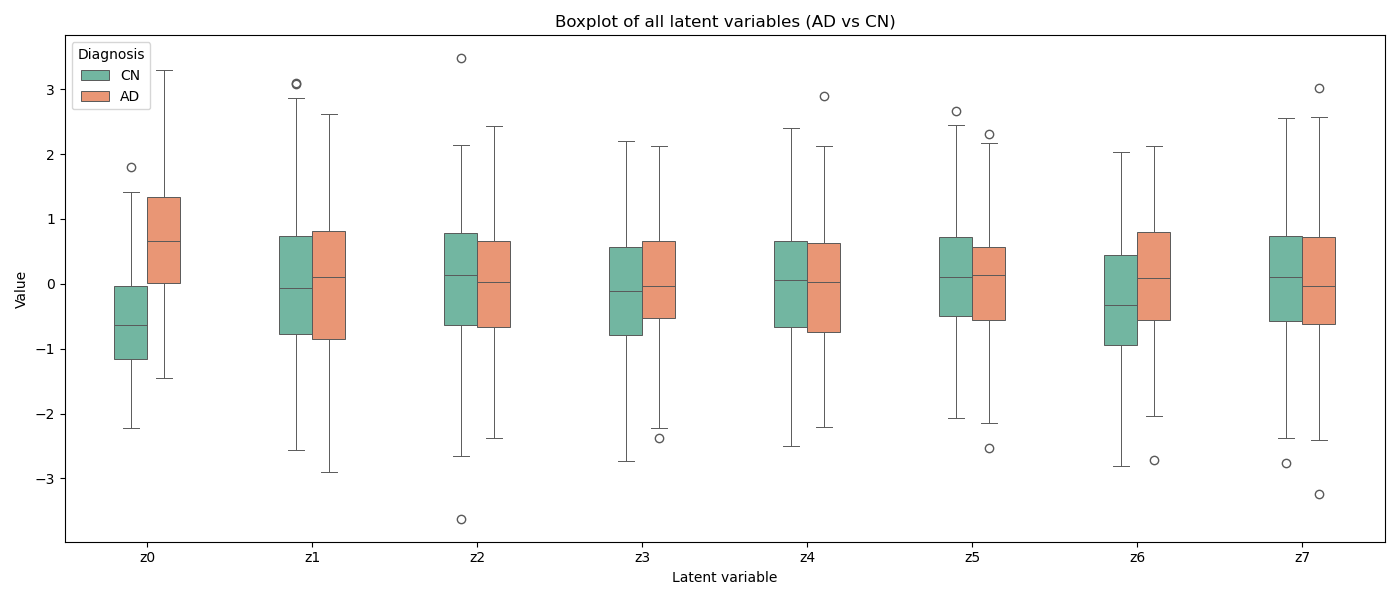}
	\caption{Classification performance (AD vs. HC). The figure shows the contribution of each latent variable ($z_{(k)}$, $k=0,\dots 7$) to the prediction.}
	\label{fig:classif_latents}
	\end{figure}

\begin{figure}[ht]
		\centering
		\includegraphics[width=0.4\textwidth]{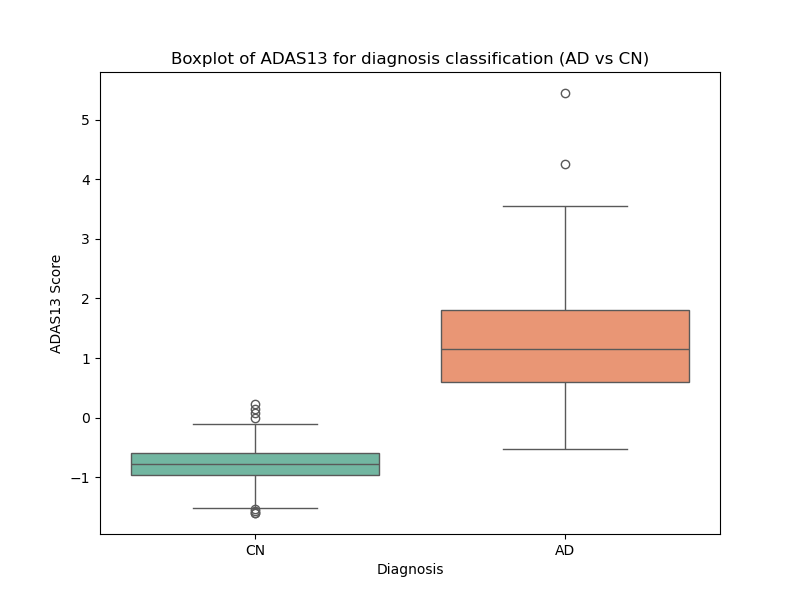}
		\caption{Classification based on the dementia score \acrshort{adas13}.}
		\label{fig:class_adas}
	\end{figure}

	Additionally, in Fig. \ref{fig:classif_latents} we show the individual contribution of each latent variable to the classification task. Here, we visualize the contribution of each dimension separately through boxplots. We find that the similarity-related latent variable $z_0$ accounts for most of the discriminative power. The remaining latent variables ($z_{(k)}$, $k=1,\dots 7$) show limited contribution. For comparison, we also include the classification performance obtained using the cognitive score (\acrshort{adas13}) as the only input (Fig. \ref{fig:class_adas}). As expected, \acrshort{adas13} achieves high predictive performance, since it serves as a clinical benchmark for disease diagnosis.

	\subsection{Disentangling age from $z_0$ \textemdash Analysis of $z_0$ and $z_1$ interaction}
\begin{figure}[H]
		\centering
		
		\begin{subfigure}[b]{0.45\linewidth}
 			\includegraphics[width=\linewidth]{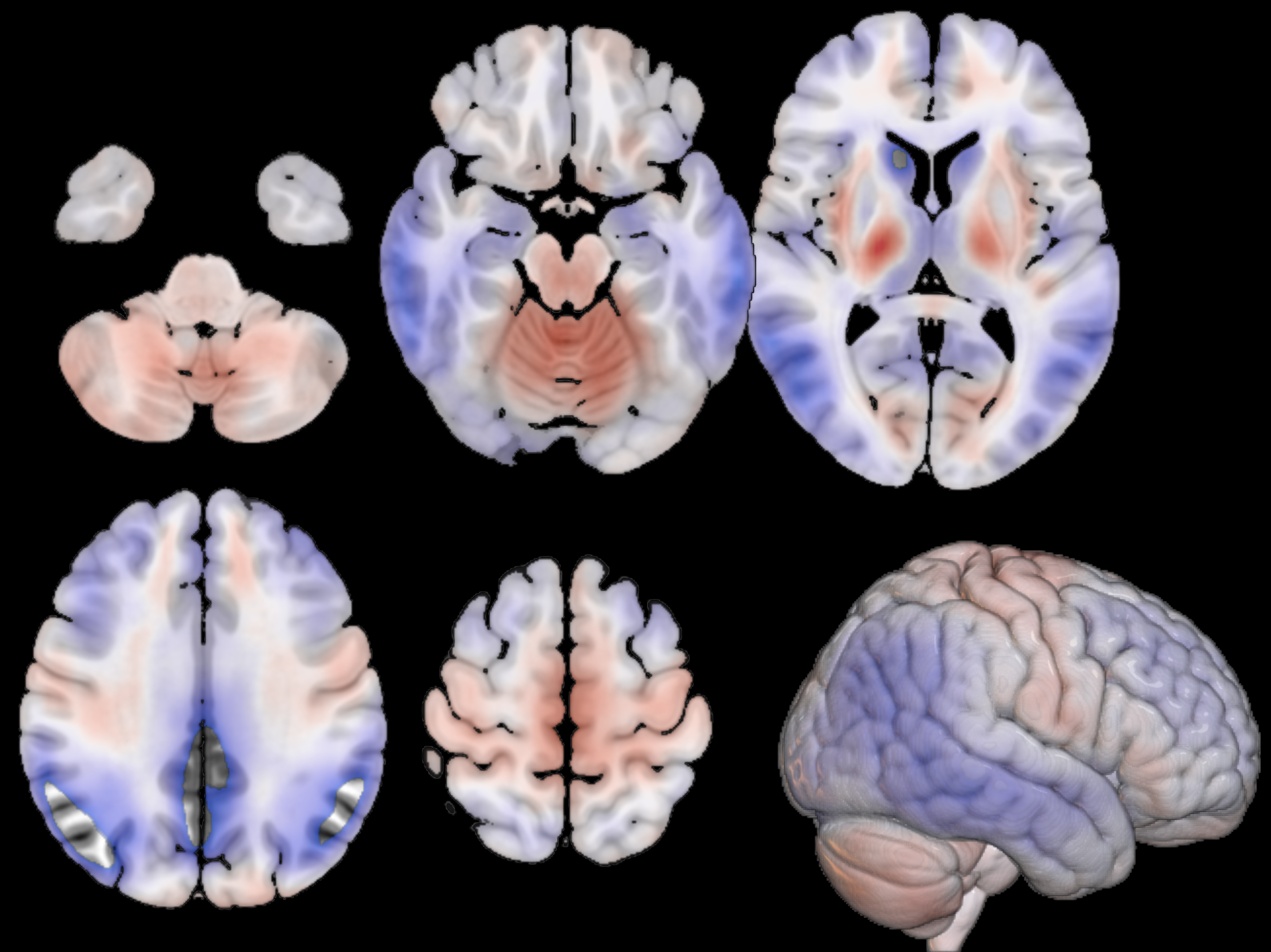}
			\caption{Mosaic axial visualization of the GLM coefficientes $\beta_0[i,j,k]$ of $z_0$ effect conditioned to $z_1$ and their interaction.}
			\label{fig:z0_effect_glm}
		\end{subfigure}
		\hfill
		\begin{subfigure}[b]{0.45\linewidth}
			\includegraphics[width=\linewidth]{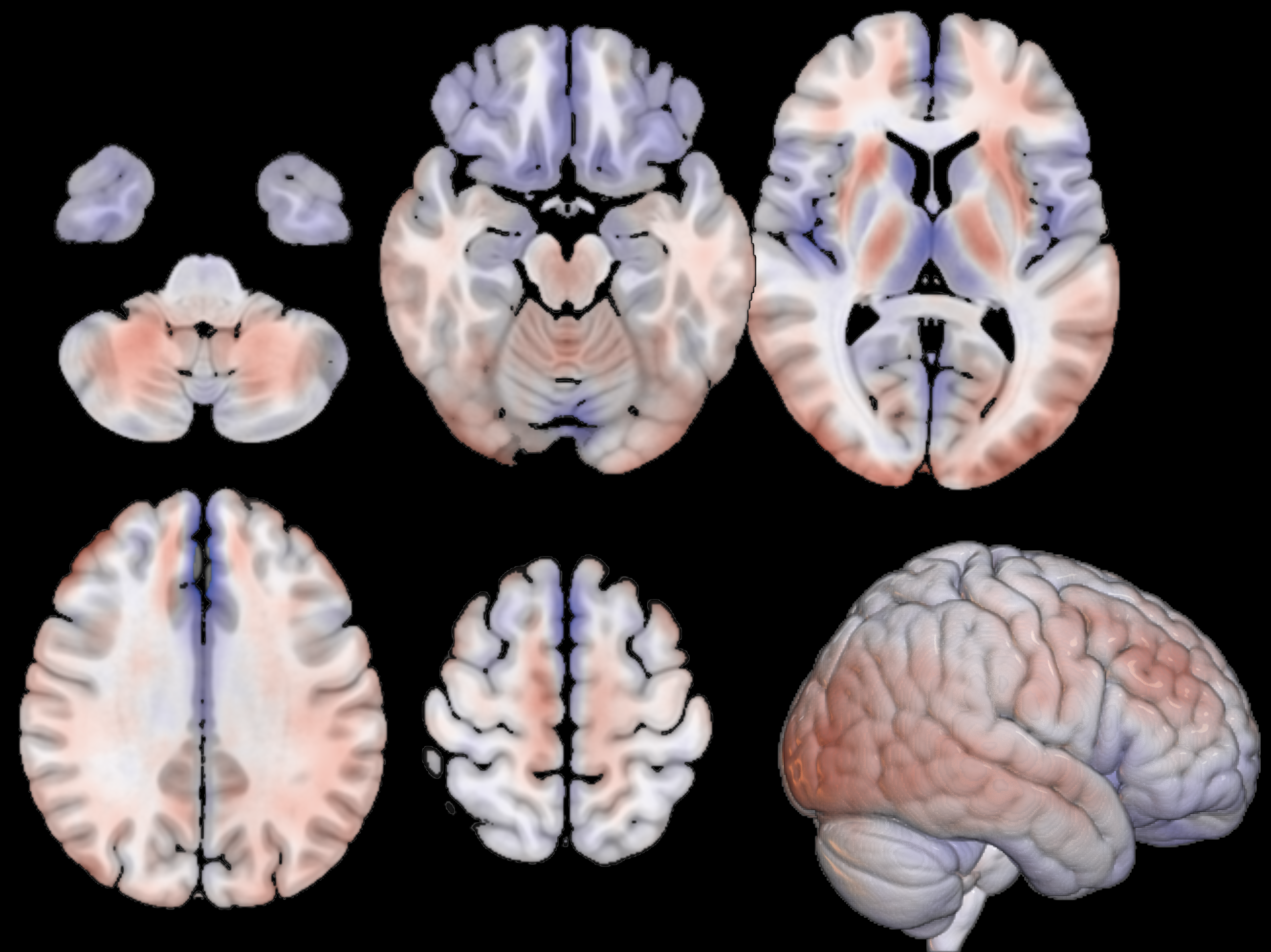}
			\caption{Mosaic axial visualization of the GLM coefficientes $\beta_1[i,j,k]$ of $z_1$ effect conditioned to $z_0$ and their interaction.}
			\label{fig:z1_effect_glm}
		\end{subfigure}
		
		\vspace{1em}
		
		\begin{subfigure}[b]{0.45\linewidth}
			\centering
			\includegraphics[width=\linewidth]{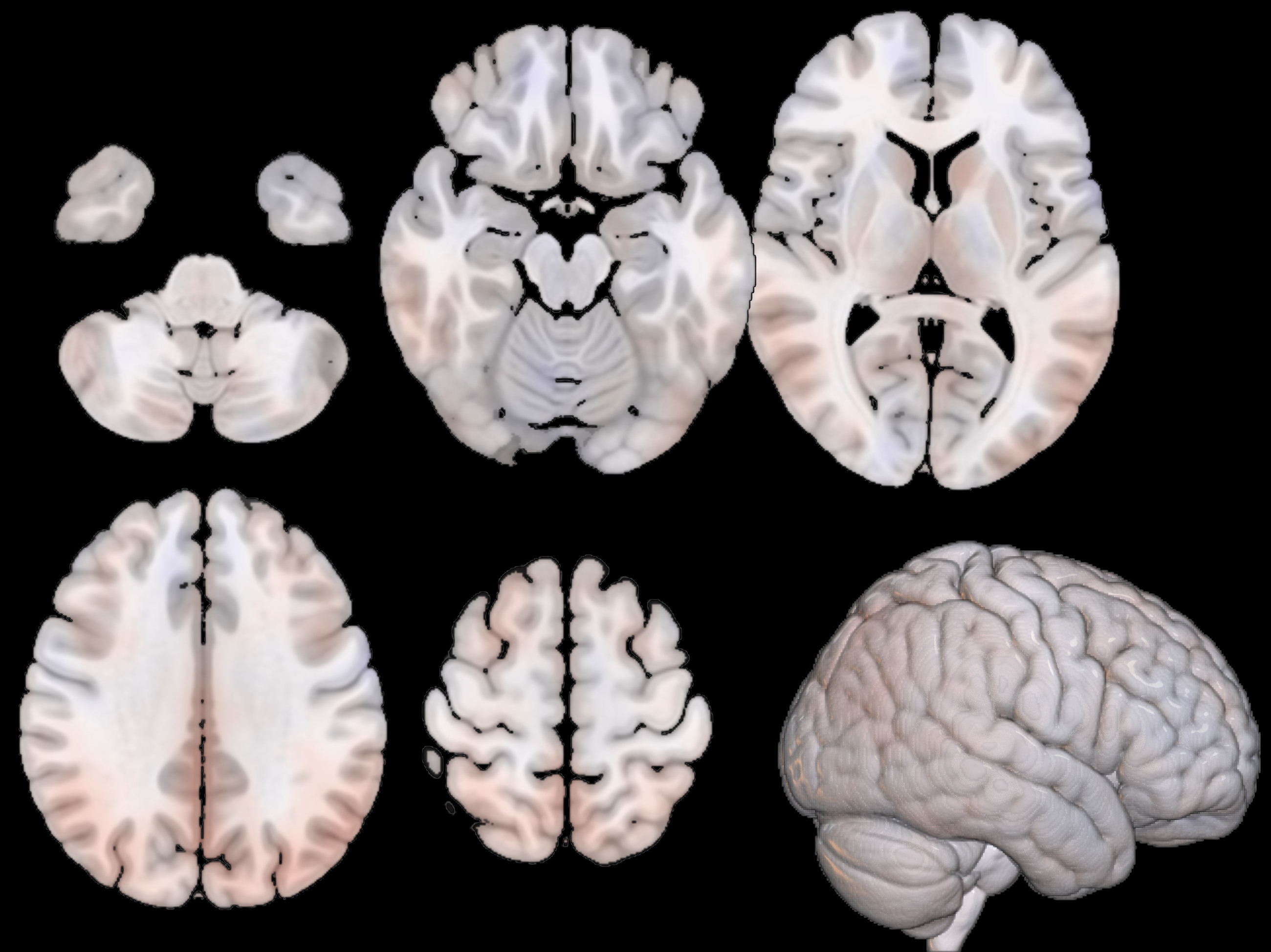}
			\caption{Mosaic axial visualization of the GLM coefficientes $\gamma_{0,1}[i,j,k]$ of the interaction between $z_0$ and $z_1$.}
			\label{fig:interaction_glm}
		\end{subfigure}
		
		\caption{Voxel-wise coefficients are shown for (a) disease severity conditioned on age and their interaction, (b) age conditioned on severity and their interaction, and (c) their direct interaction. These results indicate that the interaction between dementia severity and age is effectively disentangled: the coefficients $\gamma_{0,1}$ corresponding to their interaction are extremely small compared to the other \acrshort{glm} maps, suggesting that the observed variability is likely attributable to noise rather than to a meaningful effect.}
		\label{fig:z0_z1_interaction_glm}
	\end{figure}

	Up to this point, we have presented the \acrshort{glm} maps for $z_0$, although these estimates are conditioned on age and their interaction, as specified in Eq. \ref{general_glm_eq}. However, the results in Fig. \ref{fig:z0_z1_interaction_glm} show that the interaction between $z_0$ and $z_1$ is not significant. This indicates that the model effectively disentangles disease severity from age.

	Fig. \ref{fig:z0_effect_glm} shows the $\beta_0[i,j,k]$ coefficients previously presented in Fig. \ref{fig:adas_GLM}, reflecting the dementia-related patterns conditioned on $z_1$. In contrast, Fig. \ref{fig:z1_effect_glm} displays the $\beta_1[i,j,k]$ coefficients corresponding to age-related patterns, conditioned on $z_0$. The very small values of the interaction coefficients $\gamma_{0,1}$ suggest that the model effectively disentangles the effects of dementia severity and age.

	\subsection{Confounders \textemdash Exploration of the latent space}
\begin{figure}[tp]
		\centering
		
		\begin{subfigure}[b]{0.6\linewidth}
			\includegraphics[width=\linewidth]{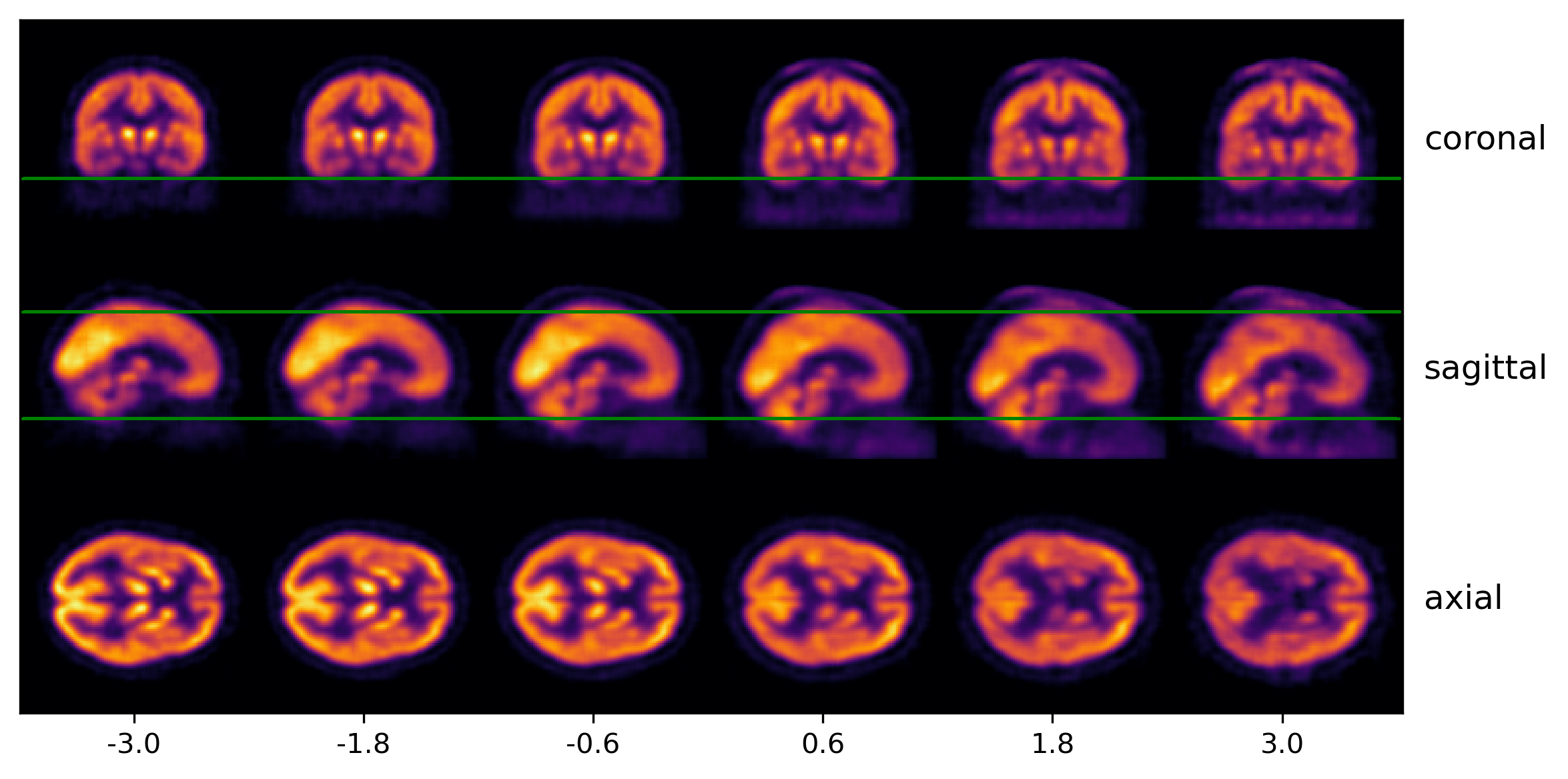}
			\caption{Manifold of latent variable 3}
			\label{fig:confounder_lat3}
		\end{subfigure}
		\hfill
		\begin{subfigure}[b]{0.6\linewidth}
			\includegraphics[width=\linewidth]{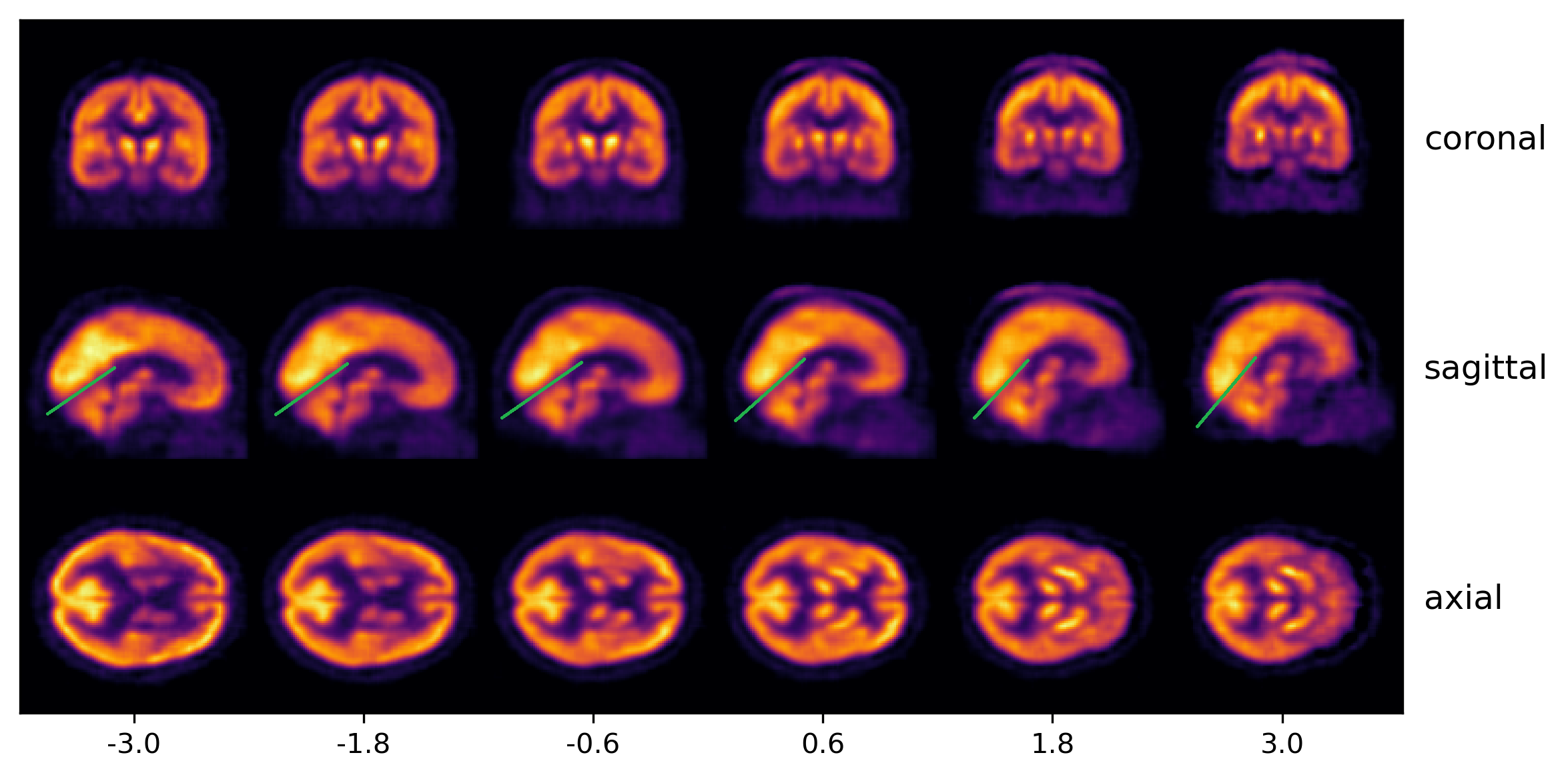}
			\caption{Manifold of latent variable 4}
			\label{fig:confounder_lat4}
		\end{subfigure}
		
		\vskip\baselineskip

		\begin{subfigure}[b]{0.6\linewidth}
			\includegraphics[width=\linewidth]{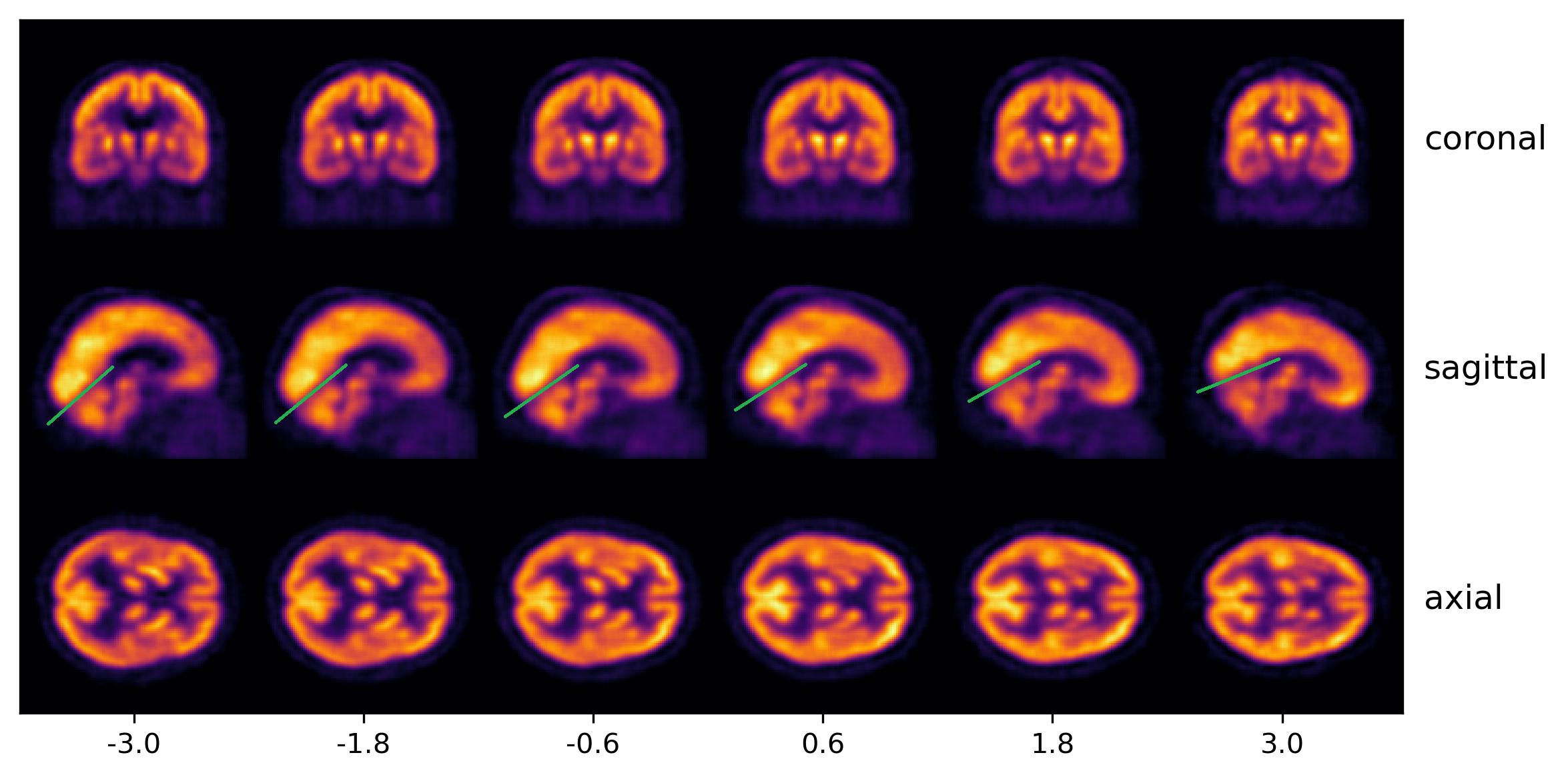}
			\caption{Manifold of latent variable 5}
			\label{fig:confounder_lat5}
		\end{subfigure}
		\hfill
		\begin{subfigure}[b]{0.6\linewidth}
			\includegraphics[width=\linewidth]{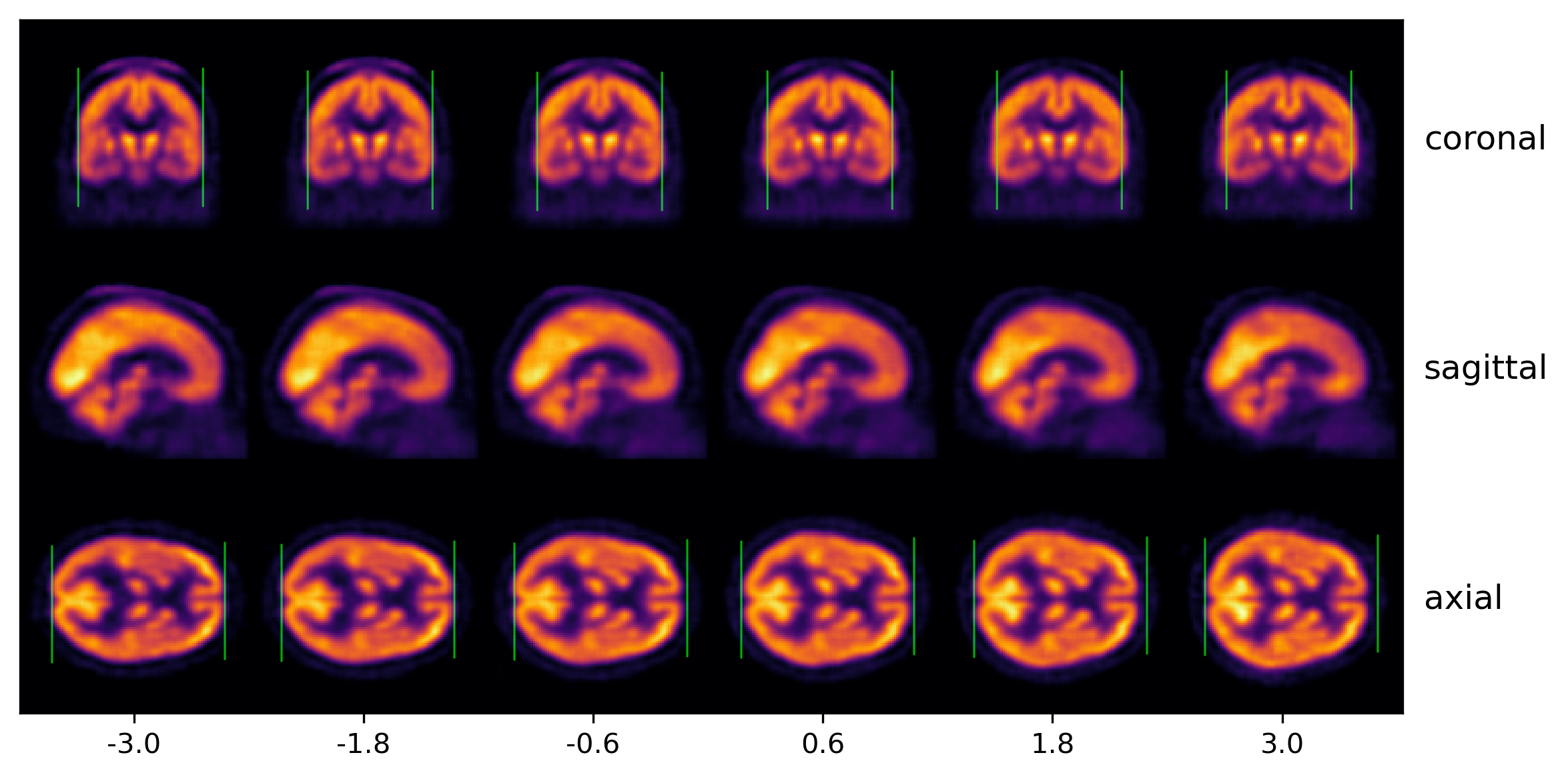}
			\caption{Manifold of latent variable 7}
			\label{fig:confounder_lat7}
		\end{subfigure}
		
		\caption{Reconstruction of the manifold for different latent variables. We observe that affine transformations are encoded within the confounder variables. (a) depicts a translation along the Z-axis, clearly visible in the sagittal plane. (b) and (c) show counterclockwise and clockwise rotations, respectively, also observable in the sagittal plane and highlighted by the green line over the tentoriym cerebelli region. (d) illustrates a scaling effect, where elongation occurs along both the X and Y axis.}
		\label{fig:confounders}
	\end{figure}

	In order to understand how other biological and noise factors are incorporated within this framework, we now explored the remaining variables of the latent space. In fig. \ref{fig:confounder_lat3}, we show the effect of modifying latent variable 3: positive values lead to a downward displacement along the Z-axis, whereas negative values correspond to upward shifts.
	
	Similarly, Fig. \ref{fig:confounder_lat5} demonstrates that latent variable 5 encodes a rotation along the Y axis. This effect is particularly visible in the orientation of the tentorium cerebelli\textemdash the membrane that separates the cerebrum from the cerebellum in the occipital lobe\textemdash, which rotates clockwise as the latent value increases. An analogous but opposite transformation is observed for latent variable 4 (Fig. \ref{fig:confounder_lat4}), which induces a counterclockwise rotation in the same axis, again discernible through the position of the tentorium cerebelli.
	
	In contrast, latent variable 7 (Fig. \ref{fig:confounder_lat7}) appears to encode a shape transformation. Negative values of this variable produce more elongated brains, whereas positive values result in shorter, more compact morphologies.
	
	Overall, we find that most latent dimensions predominantly encode affine transformations, including translations (e.g., Fig \ref{fig:confounder_lat3}), rotations (e.g., Figs. \ref{fig:confounder_lat4} and \ref{fig:confounder_lat5}), and scaling effects (e.g., \ref{fig:confounder_lat7}). Furthermore, the colormap of the reconstructions indicates that the latent variables encode not only structural information, but also intensity-related variations. For example, positive values of latent variable 3 are associated with a more sharply defined skull and reduced contrast between the brain and surrounding tissue. In contrast, negative values enhance the contrast between the brain and the background structures, producing a skull-stripping effect. Additionally, we observe a general decrease in intensity in specific brain regions, which may reflect reduced tracer uptake.

	\subsection{Ablation study}
\begin{figure}[H]
		\centering
		\begin{subfigure}[b]{0.45\textwidth}
			\centering
			\includegraphics[width=\textwidth]{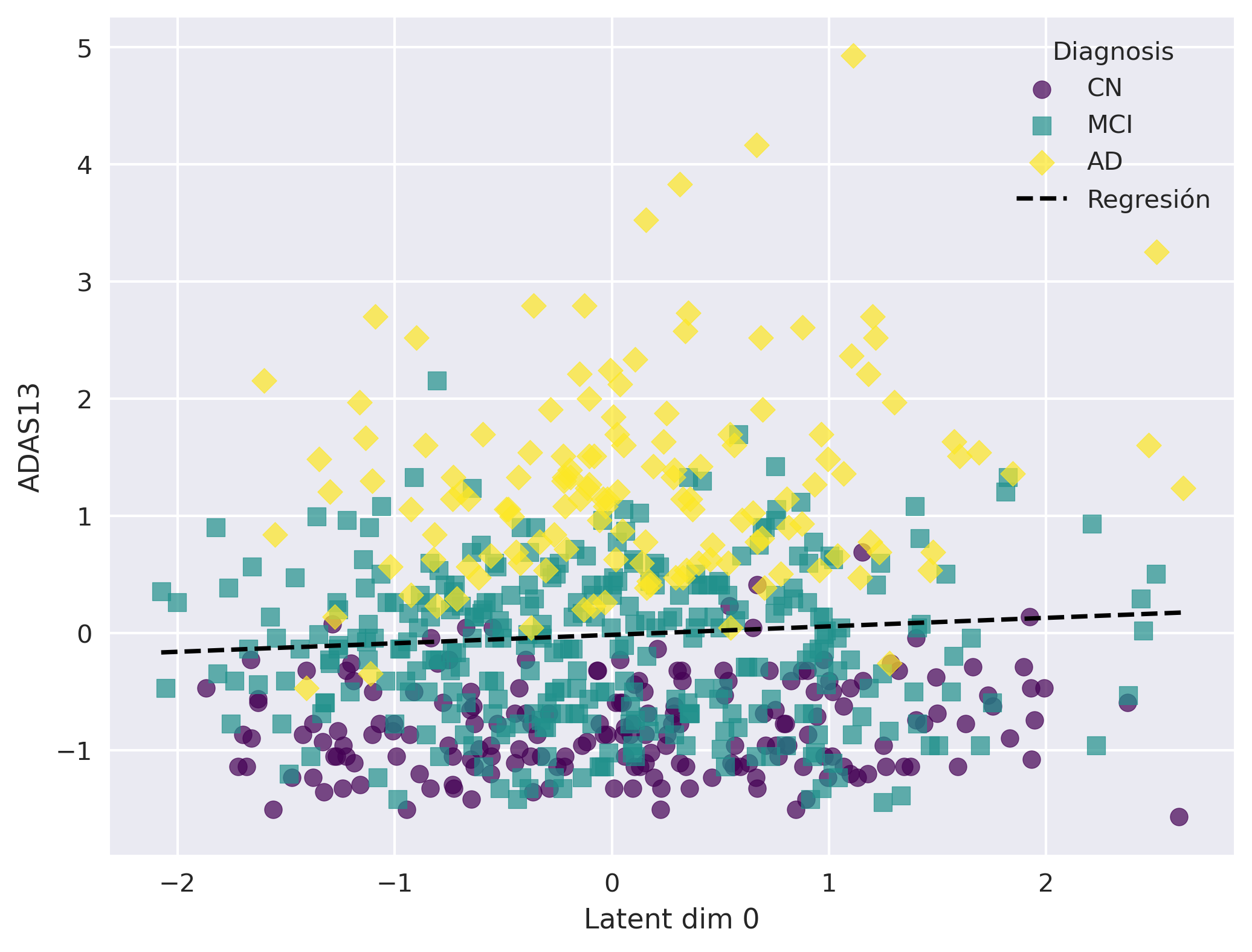}
			\caption{Scatterplot showing $z_0$ vs \acrshort{adas13} in the $\beta$-VAE model with no similarity regularization. Subjects are colored depending on their diagnosis.}
			\label{fig:ablation_adas_z0}
		\end{subfigure}
		\hfill
		\begin{subfigure}[b]{0.45\textwidth}
			\centering
			\includegraphics[width=\textwidth]{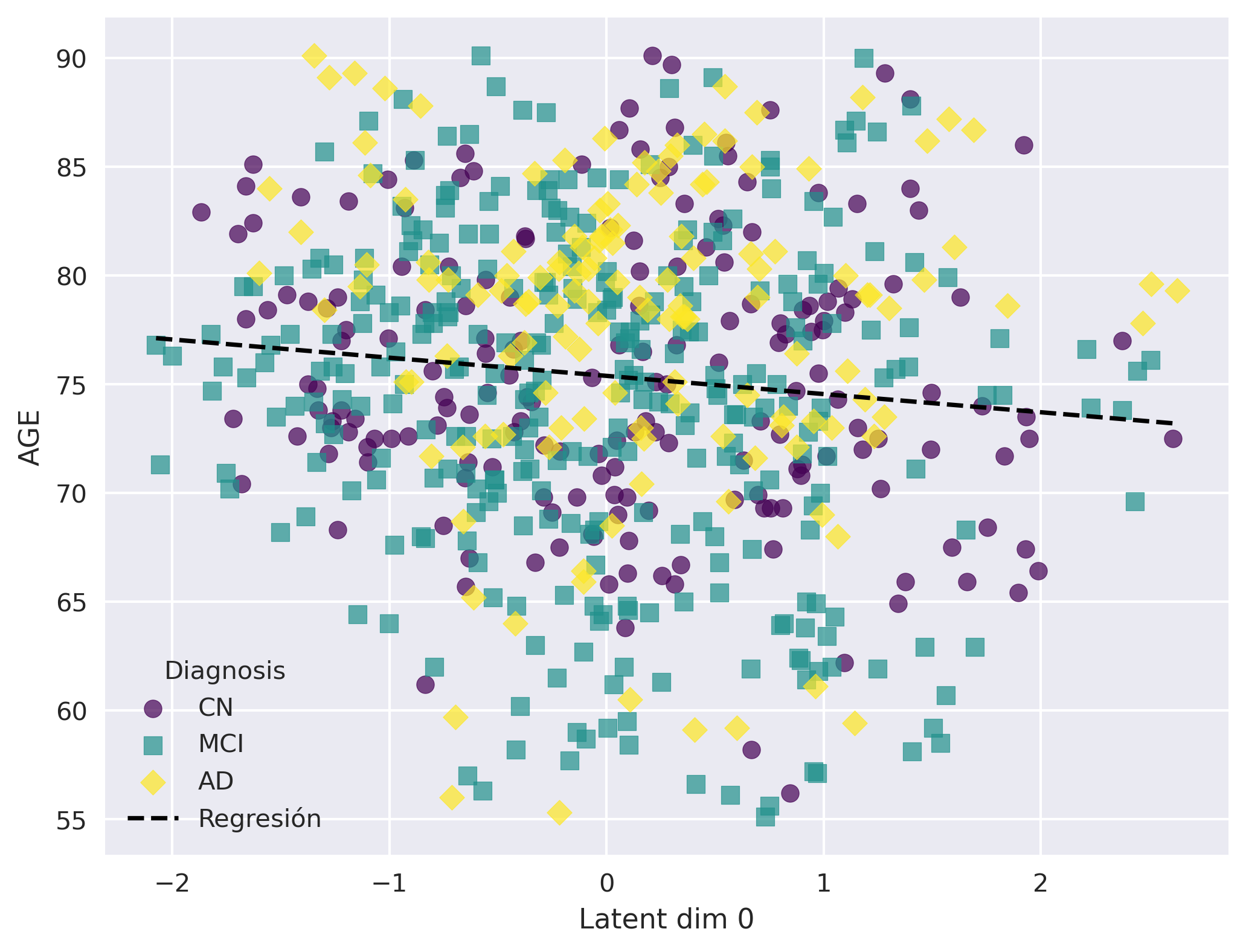}
			\caption{Scatterplot showing $z_1$ vs age in the $\beta$-VAE model with no similarity regularization. Subjects are colored depending on their diagnosis.}
			\label{fig_ablation_age_z1}
		\end{subfigure}
		\caption{}
		\label{fig:ablation}
	\end{figure}

	We conducted an ablation study by removing the similarity regularization term \eqref{similarity_reg}, thereby reducing the model to a standard $\beta$-\acrshort{vae}. Fig. \ref{fig:ablation} illustrates the correlation between $z_0$ and \acrshort{adas13}, and between $z_1$ and age, respectively. These results are presented in direct comparison with the previous analyses of \acrshort{adas13} (see Fig. \ref{fig:corr_adas}), but no correlations are found between the latent variables and any biological feature, such as temporal lobes, ventricles, etc.
	
\begin{figure}[H]
		\centering
		\begin{subfigure}[b]{0.41\textwidth}
			\centering
			\includegraphics[width=\textwidth]{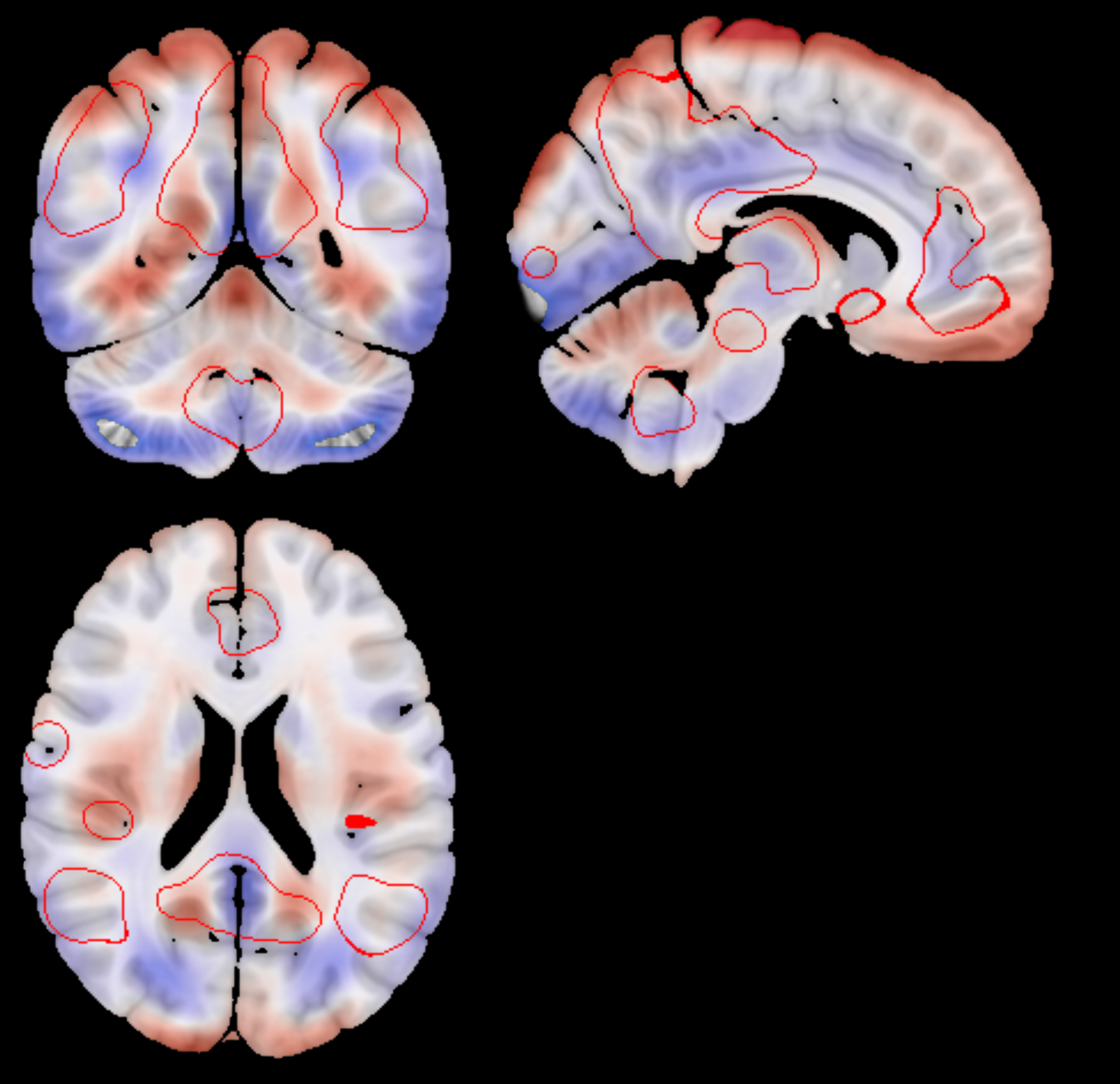}
			\caption{Visualization of the \acrshort{dmn}, highlighted with red outlines.}
			\label{fig:glm_dmn}
		\end{subfigure}
		\hfill
		\begin{subfigure}[b]{0.45\textwidth}
			\centering
			\includegraphics[width=\textwidth]{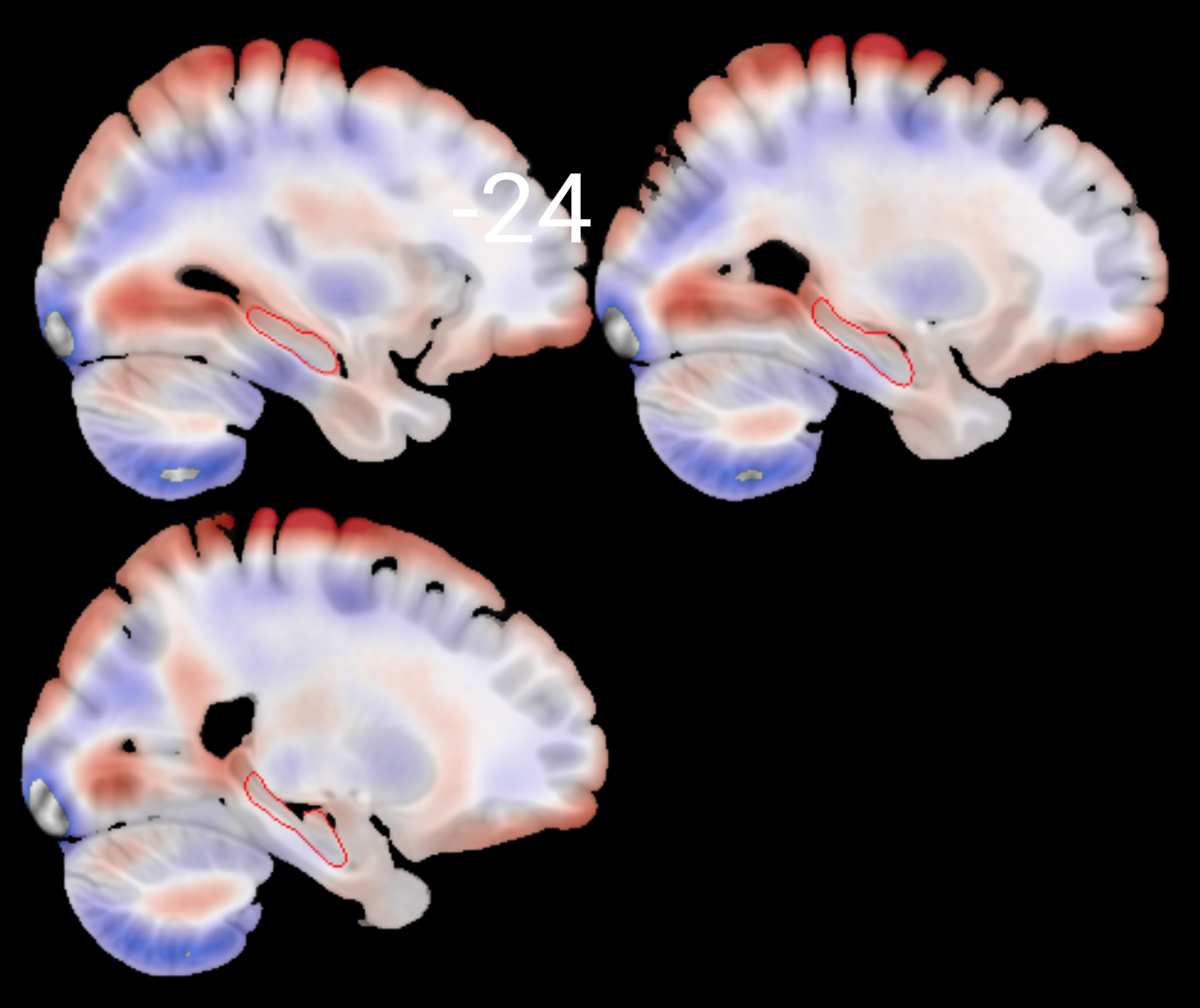}
			\caption{Visualization of the hippocampus, highlighted with red outlines.}
			\label{fig:glm_hipo}
		\end{subfigure}
		\caption{Voxel coefficients for $z_0$ estimated by the \acrshort{glm} model for the ablated model with no similarity loss. Values correspond to $\beta_0[i,j,k]$ coefficients of the model. Direct comparison with Fig. \ref{fig:adas_GLM} reveals that no decline in relevant structures, such as hippocampus and \acrshort{dmn} arise from this model. }
		\label{fig:ablation_glm}
	\end{figure}

	A complementary comparison is shown in Fig. \ref{fig:ablation_glm}. The corresponding \acrshort{glm} maps indicate the absence of significant correlations with key structures such as the \acrshort{dmn} and the hippocampus. This contrasts with the findings obtained when the similarity regularization term is included in the model (see Fig. \ref{fig:adas_GLM}). The same behavior is observed for the rest of the variables.
	\acrshort{adas13} \ref{fig:corr_adas}.


	\section{Discussion}
	
	 In this study, we developed a semi-supervised \acrshort{vae} to learn an interpretable neuroimaging biomarker of dementia severity from \acrshort{fdg} scans of the \acrshort{adni} database. Our primary objective was to derive a latent variable $z_0$ that encodes disease progression patterns and correlates with clinical measures, while disentangling confounding factors into separate latent dimensions.
	To encourage the model to learn informative latent representations, we introduced a regularization strategy based on a set of similarity losses \textemdash \eqref{similarity_reg} controlled by the hyperparameters $\alpha_j$\textemdash that guide a set of latent variables to capture features related to dementia severity, and confounding factors. In particular, to conduct the experiments, we chose the similarity function to be the Pearson correlation, defined by \eqref{pearson_formula}, and the similarity terms to be the correlations between ($z_0, \text{ADAS13}$) and ($z_1, \text{age}$). However, we also demonstrated that the choice of similarity metric is flexible, as retraining the model using Spearman correlation still yielded high correlation values.
	
	First, we conducted an exploratory study to ensure the model was effectively utilizing the latent space. The resulting figures \ref{fig:phase_diagram_conv} and \ref{fig:phase_diagram_corr} provide us with a proxy for selecting hyperparameters that avoid posterior collapse to the mean and provide informative representations of neurodegeneration progression. This analysis is crucial for the study of \acrshort{vae}s in neuroimaging, as complex data is sensitive to hyperparameters, as shown in the figures. Moreover, because no analytical method currently exists to determine optimal hyperparameter values, such empirical evaluation becomes essential \cite{ichikawa2023dataset}.
	
	Next, we selected a fixed set of hyperparameters and studied the resulting latent representations of the model, demonstrating the model's ability to encode dementia patterns that might otherwise go unnoticed in neuroimaging data (Fig. \ref{fig:corrs_z0}). As shown in Fig. \ref{fig:biomarkers}, several key biomarkers associated with \acrshort{ad} exhibit correlation with the dementia-related variable. Notably, hippocampal volume and medial temporal lobe volume demonstrate a particularly strong association. In contrast, the fusiform gyrus shows only a weak correlation, which is consistent with its involvement in later stages of the disease \cite{braak1991neuropathological}.
	
	The generative nature of the \acrshort{vae} allowed us to map these dementia patterns back to brain space, providing an interpretable framework for visualizing the patterns learned by the model to distinguish between \acrshort{ad} subjects and normal controls. By performing a GLM on reconstructions of subjects with different values of the $z_0$ and $z_1$ latent variables, we were able to visualize the variations due to dementia in the brain space (Fig. \ref{fig:adas_GLM}, using the $\beta_0[i,j,k]$ coefficients from \eqref{general_glm_eq}). We found that the patterns learned by the model correspond to well-known regions affected by the disease. A marked reduction in glucose metabolism is observed within the \acrshort{dmn} \textemdash a well-established hallmark of AD \cite{greicius2004default, buckner2008brain}, as seen in Fig. \ref{fig:adas_GLM_DMN}. Similarly, Figs. \ref{fig:adas_GLM_FPN_left} and \ref{fig:adas_GLM_FPN_right} show that the \acrshort{cen}, which encompasses both \acrshort{fpn}s, also shows reduced metabolic activity, in line with previous literature. The \acrshort{dmn} is involved in internally directed cognitive processes such as introspection and memory retrieval, while the \acrshort{cen} supports functions like attention, working memory, and cognitive control \cite{raichle2001default, seeley2007dissociable}. These domains are significantly impaired in \acrshort{ad} \cite{sorg2007selective, zhou2010divergent, brier2014network}, making the observed metabolic patterns both biologically plausible and interpretable.
	
	In contrast, the Sensorimotor Network \textemdash typically preserved in \acrshort{ad}\textemdash shows either a slight increase or no significant change in metabolic activity. These patterns are clearly localized by our model mainly to the precentral and postcentral gyri, areas responsible for voluntary motor control and somatosensory processing \cite{mosconi2005brain, sperling2010functional}. We observe a similar behavior in the central occipital cortex, which is also known to be unaffected by \acrshort{ad} \cite{scahill2002mapping}.
	Moreover, analyzing the interaction between $z_0$ and $z_1$ we found that, interestingly, while \acrshort{adas13} inherently reflects both disease-related cognitive decline and age-related effects, our analyses indicate that the dementia-related latent variable $z_0$ is effectively disentangled from age. As shown in Fig. \ref{fig:z0_z1_interaction_glm}, the voxel-wise coefficients for the interaction between $z_0$ and $z_1$ (age) are extremely small, suggesting that age does not meaningfully influence the disease-related latent representation.
	
	Furthermore, we found that the remaining variables in the latent space offer an excellent framework for understanding how typical confounders, arising from both subject variability and acquisition noise, are integrated into the model. This significantly contributes to a more explainable and robust model, enabling the disentanglement and characterization of common confounding factors \textemdash such as brain size or orientation\textemdash within the latent representation, treating them as informative components rather than as noise \cite{alfaro2021confound}. As illustrated in Fig. \ref{fig:confounders}, affine transformations are encoded within almost every latent variable, accounting for translations (\ref{fig:confounder_lat3}), rotations (\ref{fig:confounder_lat4},\ref{fig:confounder_lat5}), and scaling (\ref{fig:confounder_lat7}) due to varying brain shapes and positions during acquisition. Moreover, we also found that intensity variations are encoded within these variables. We observe that some variables (e.g., latent variable 3) display a skull-stripping effect. On the other hand, we also observe a general decrease in intensity for specific brain areas. This variability could be attributable to factors such as differences in acquisition timing (e.g., delayed imaging leading to lower uptake) or variations in administered dose, which can differ between subjects in clinical PET protocols \cite{murthy202068, doot2010instrumentation}. Although intensity normalization is applied to mitigate such inter-subject variability, residual effects might still persist, potentially impacting the observed signal in specific regions. 
	
	Finally, although not the primary objective of this work we also performed a logistic regression trained on the learned representations, demonstrating that the $z_0$ latent captures substantial disease-related information, validating its biological relevance, as it is capable of discriminating \acrshort{hc} from \acrshort{ad}. Our model achieved excellent values of the metrics. We found that the performance, shown in Table \ref{tab:classification_report} and Fig. \ref{fig:classif_latents} is comparable to the previous studies \cite{wakefield2024variational, bit2024mri, vivek2023explainable}. Furthermore, the dementia-related variable explains the majority of the classification's predictive power, whereas the remaining latent variables perform poorly, indicating that the \acrshort{vae} model primarily encodes dementia-related information within the similarity-related variable.
	
	Moreover, an ablation study in which the similarity regularization term \eqref{similarity_reg} (responsible for guiding the latent representations toward disease-relevant patterns) was removed demonstrates its critical role in the model. In the absence of this term, no significant correlations (Fig. \ref{fig:ablation_adas_z0}) or meaningful spatial patterns (Fig. \ref{fig:ablation_glm}) are observed. These findings indicate that the similarity regularization is essential for the model to capture clinically relevant structure.
	
	Overall, this work presents a flexible and generalizable semi-supervised VAE framework that effectively integrates clinical variables through a similarity-driven regularization. By incorporating noise as part of the latent representation, the model not only captures meaningful disease-related patterns but also accounts for subject variability in an interpretable way. Moreover, results have shown that this neuroimaging biomarker is also disentangled from age.
	Our approach provides the derivation of an interpretable latent biomarker directly from neuroimaging data, capturing the underlying progression of dementia. This opens remarkable possibilities for advancing diagnosis, monitoring, and our overall understanding of the disease.
	
	\section*{Ethics statement}
	
	Data collection and sharing in ADNI was approved by the Institutional Review Board of each participating institution. Written informed consent was obtained from all ADNI participants and/or authorized representatives before any protocol-specific procedures were carried out. All procedures involving human participants were in accordance with the ethical standards of the 1964 Helsinki Declaration and its later amendments.
	
	\section*{Acknowledgments}
	This publication is part of the projects PID2022-137629OA-I00 and PID2022-137451OB-I00 funded by MICIU/AEI/10.13039/501100011033 and by ERDF/EU, and the C-ING-183-UGR23 project, cofunded by the Consejería de Universidad, Investigación e Innovación and by European Union, funded by Programa FEDER Andalucía 2021-2027. F.J.M.M. received support from grant RYC2021-030875-I, funded by MICIU/AEI/\-10.13039/\-501100011033 and European Union NextGenerationEU/PRTR.
	
	Data collection and sharing for the \acrshort{adni} is funded by the National Institute on Aging (National Institutes of Health Grant U19AG024904). The grantee organization is the Northern California Institute for Research and Education. In the past, \acrshort{adni} has also received funding from the National Institute of Biomedical Imaging and Bioengineering, the Canadian Institutes of Health Research, and private sector contributions through the Foundation for the National Institutes of Health (FNIH) including generous contributions from the following: AbbVie, Alzheimer's Association; Alzheimer's Drug Discovery Foundation; Araclon Biotech; BioClinica, Inc.; Biogen; Bristol-Myers Squibb Company; CereSpir, Inc.; Cogstate; Eisai Inc.; Elan Pharmaceuticals, Inc.; Eli Lilly and Company; EuroImmun; F. Hoffmann-La Roche Ltd.; Janssen Alzheimer Immunotherapy Research $\&$ Development, LLC.; Johnson $\&$ Johnson Pharmaceutical Research $\&$ Development LLC.; Luminosity; Lundbeck; Merck $\&$ Co., Inc.; Meso Scale Diagnostics, LLC.; NeuroRx Research; Neurotrack Technologies; Novartis Pharmaceuticals Corporation; Pfizer Inc.; Piramal Imaging; Servier; Takeda Pharmaceutical Company; and Transition Therapeutics.

	
	\bibliographystyle{elsarticle-harv} 
	\bibliography{refs_review.bib}
	
	
	
	
	
\end{document}